# Application of deep learning for livestock behaviour recognition: A systematic literature review

Ali Rohan[a,e,*], Muhammad Saad Rafaq[b], Md. Junayed Hasan[a], Furqan Asghar[c], Ali Kashif Bashir[d] and Tania Dottorini[e,*]

[a]*National Subsea Centre, School of Computing, Robert Gordon University, 3 International Ave, Dyce, Aberdeen, AB21 0BH, UK*
[b]*Wolfson School of MEME, Loughborough University, Loughborough, LE11 3TU, UK*
[c]*Department of Energy Systems Engineering, University of Agriculture, Faisalabad, 38000, Pakistan*
[d]*Department of Computing and Mathematics, Manchester Metropolitan University, Manchester, M15 6BY, UK*
[e]*School of Veterinary Medicine and Sciences, University of Nottingham, Sutton Bonington Campus, Loughborough, LE12 5RD, UK*

ABSTRACT

Livestock health and welfare monitoring is a tedious and labour-intensive task previously performed manually by humans. However, with recent technological advancements, the livestock industry has adopted the latest AI and computer vision-based techniques empowered by deep learning (DL) models that, at the core, act as decision-making tools. These models have previously been used to address several issues, including individual animal identification, tracking animal movement, body part recognition, and species classification. However, over the past decade, there has been a growing interest in using these models to examine the relationship between livestock behaviour and associated health problems. Several DL-based methodologies have been developed for livestock behaviour recognition, necessitating surveying and synthesising state-of-the-art. Previously, review studies were conducted in a very generic manner and did not focus on a specific problem, such as behaviour recognition. To the best of our knowledge, there is currently no review study that focuses on the use of DL specifically for livestock behaviour recognition. As a result, this systematic literature review (SLR) is being carried out. The review was performed by initially searching several popular electronic databases, resulting in 1101 publications. Further assessed through the defined selection criteria, 126 publications were shortlisted. These publications were filtered using quality criteria that resulted in the selection of 44 high-quality primary studies, which were analysed to extract the data to answer the defined research questions. According to the results, DL solved 13 behaviour recognition problems involving 44 different behaviour classes. 23 DL models and 24 networks were employed, with CNN, Faster R-CNN, YOLOv5, and YOLOv4 being the most common models, and VGG16, CSPDarknet53, GoogLeNet, ResNet101, and ResNet50 being the most popular networks. Ten different matrices were utilised for performance evaluation, with precision and accuracy being the most commonly used. Occlusion and adhesion, data imbalance, and the complex livestock environment were the most prominent challenges reported by the primary studies. Finally, potential solutions and research directions were discussed in this SLR study to aid in developing autonomous livestock behaviour recognition systems.

## 1. Introduction

The livestock farming sector is crucial in the global food system, significantly contributing to food security, agriculture development, and poverty reduction. According to the Food and Agriculture Organization (FAO) of the United Nations (UN), livestock contributes 40% of the global agricultural output, with 1.3 billion people depending on livestock for their livelihood, food, and nutrition security (Food and Agriculture Organization of the United Nations 2022). Livestock farming involves managing and breeding domestic animals, such as cattle, pigs, and poultry, to produce meat, eggs, milk, and other dairy products for human consumption. With population growth and changing dietary patterns, the demand for livestock farming has surged in recent decades, making it one of the fastest-growing agricultural sub-sectors. However, this rapid growth comes at a cost. It is predicted that if the growth of the livestock sector is not managed efficaciously, environmental sustainability and public health issues will become more complicated (Nowodziński, 2021). Therefore, it is critical to adopt sustainable livestock farming practices to ensure that future generations can enjoy the benefits of this sector.

In livestock farming, studying animal behaviour is crucial in understanding how animals interpret and respond to their environment. This enables us to use effective techniques to improve their health and welfare on farms. Although animal behaviour research has been conducted for many years, it is only recently that applying animal behaviour to production, health, and welfare has become more popular (Orihuela, 2021). Identifying certain animal behaviours helps detect any potential underlying health problems. For instance, an abnormal gait pattern in cows might indicate lameness (Flower et al., 2005), (Flower et al., 2007). Moreover, the amount of time spent in feeding, feeding frequency, and amount of food ingested can help differentiate between lame cows and non-lame cows (Thorup et al., 2016). Health disorders like ketosis and mastitis can decrease feed intake and milk production (Bareille et al., 2003). The nursing behaviour of sows is closely linked to piglet death caused by starvation (Shankar et al., 2009). Research has shown that dangerous body movements of sows can increase the mortality rate of piglets (Chidgey et al., 2015). Early recognition of animal behaviours is crucial for optimising animal production processes, improving animal health and welfare, increasing the quality and productivity of animal products, and preventing losses caused by diseases (Nasirahmadi et al., 2017).

*Corresponding authors at National Subsea Centre, School of Computing, Robert Gordon University, Aberdeen AB10 7AQ, UK (A.R.); School of Veterinary Medicine and Sciences, University of Nottingham, Sutton Bonington Campus, Loughborough, LE12 5RD, UK (T.D.). E-mail addresses: a.rohan@rgu.ac.uk (A.R.), tania.dottorini@nottingham.ac.uk (T.D).

**Table 1**
Nomenclature

| Nomenclature | |
|---|---|
| BiFPN | Bi-directional Feature Pyramid Network |
| Bi-LSTM | Bi-directional Long Short-Term Memory |
| C3D | Convolutional 3D |
| ConvLSTM | Convolutional Long Short-Term Memory |
| CNN | Convolutional Neural Network |
| LSTM | Long Short-Term Memory |
| CSPDarknet | Cross Stage Partial network based Darknet |
| DHRN | Deep High Resolution Network |
| DNN | Deep Neural Network |
| DRNet | Dense Residual Network |
| DRN-YOLO | DenseResNet-You Only Look Once |
| ESCMobileNet | Extremely Separated Convolution-MobileNet |
| R-CNN | Region-based Convolutional Neural Network |
| FCN | Fully Convolutional Network |
| FD-CNN | Frame Differences-Convolutional Neural Network |
| FPN | Feature Pyramid Network |
| ResNet | Residual neural Network |
| R-FCN | Region-based Fully Convolutional Network |
| RNN | Recurrent Neural Network |
| RPN | Region Proposal Network |
| SAE | Sparse Autoencoder |
| SSD | Single-shot detector |
| STAGCN | Spatiotemporal and Adaptive Graph Convolutional Network |
| TSN | Temporal Segment Network |
| VGG | Visual Geometry Group |
| YOLO | You Only Look Once |
| ZFNet | Zeiler & Fergus Net |
| mAP | mean Average Precision |
| AP | Average Precision |
| IoU | Intersection over Union |
| MCC | Matthew's Correlation Coefficient |
| MPA | Mean Pixel Accuracy |

In traditional livestock farms, experienced workers, farmers, or veterinary behaviourists rely on their knowledge and skills to recognise animal behaviour and its association with underlying health conditions. However, as the environment of livestock farms is quite complex, where different species of animals are exploited in large groups housed in intensive production systems, it is unrealistic to observe every activity of animals in real time. With the advent of Industry 4.0 technologies in industrial automation, creating systems that can perform these tasks more efficiently and autonomously has become more practical. Advanced technologies such as Artificial Intelligence (AI) have reshaped industrial livestock farming and given rise to the fields of "Smart farming," "Precision Agriculture," and "Precision Livestock Farming" (Gonçalves et al., 2022). These fields heavily utilise AI to analyse data and provide new tools for monitoring and managing animal behaviour and health. These new technologies have the potential to revolutionise traditional livestock farming, enabling farmers to make data-driven decisions that improve animal health, welfare, and productivity. Precision livestock farming can help optimise the production process, reduce costs, and minimise environmental impacts by providing real-time monitoring and analysis of animal behaviour (Garcia et al., 2020). Also, integrating advanced technologies in livestock farming can transform the industry, leading to more sustainable and efficient practices that benefit both animals and farmers. Finally, these technologies can help the fight against infectious and endemic diseases, one of the major challenges in contemporary livestock farming, with repercussions on both animals (e.g. mastitis) and consumers' health (Grace, 2019).

In precision livestock farming, a subfield of AI known as Deep Learning (DL) has been widely used to address various issues, including individual animal identification (Qiao et al., 2020), body part recognition (Hu et al., 2020), face recognition (Yao et al., 2019), (Xu et al., 2022), health monitoring (Atkinson et al., 2020), animal tracking and counting (Laradji et al., 2020), race classification (Santoni et al., 2015), species classification (Tabak et al., 2019), and behaviour recognition.

DL has been preferred over other shallow and machine learning methods since it employs multi-layered deep neural networks that can learn the representations from raw data autonomously, allowing it to uncover detailed patterns and features. It is considered best for handling large and high-dimensional datasets, particularly for complex tasks such as image classification, object detection, and spatiotemporal analysis. Despite the many applications of DL, there still needs to be more understanding about using different DL models to solve specific problems. As a result, the challenges encountered in designing, developing, and deploying these models still need to be fully understood.

Recent review studies on precision livestock farming have explored sustainability, environmental, and socioeconomic aspects (Lovarelli et al., 2020), as well as machine learning applications for dairy farm management (Slob et al., 2021), and deep learning applications for precision cattle farming (Mahmud et al., 2021). While these studies provide valuable insights into the use of DL in livestock farming, they are general and need more focus on the use of DL to solve a specific problem such as behaviour recognition. The effectiveness of DL models and networks can vary significantly depending on the problem they are designed to solve. For example, a DL model created for object classification in images may perform poorly in a complex spatiotemporal task like behaviour recognition. Therefore, it is essential to research to establish the state of DL concerning specific problems in precision livestock farming. By doing so, we can better understand how to design, develop, and deploy DL models effectively, ultimately leading to better outcomes for livestock farmers and their animals. Therefore, this review aims to examine the recent trends and advancements in the use of DL for a specific problem of behaviour recognition in precision livestock farming. This review will provide a comprehensive overview of the various types of behaviour recognition problems that have been addressed using DL. In addition, it will summarise the different approaches employed for data collection, including the quantity, quality, and type of data used in these studies. Furthermore, it will discuss the different types of DL models and networks developed for behaviour recognition and how they are applied to specific problems, the performance analysis of DL models and networks, challenges reported in the literature, and potential solutions to overcome these challenges.

The details of this study are presented in the following sections. Section 2 defines the methodology adopted to conduct this review, including the review protocol, research questions, search strategy, and selection criteria for the primary studies. Section 3 consists of the results with answers to the research questions. Section 4 presents a general discussion and observations. Finally, Section 5 presents the conclusion.

## 2. Methodology
### 2.1. Review Protocol

This study follows the protocols for conducting a Systematic Literature Review (SLR) provided by (Kitchenham and Charters, 2007), entitled "Guidelines for Performing Systematic Literature Reviews in Software Engineering". Figure 1 outlines the steps involved in the process of SLR.

The SLR process was divided into three steps. The first step was planning the review, which involved determining the need for the review, developing research questions, and establishing search strategies. Search strategies focus on selecting relevant databases, compiling appropriate search strings, and defining selection criteria.

The second step was executing the review, which involved selecting primary studies and extracting and synthesising data. Search strings were used to search the selected databases' titles, abstracts, and keyword fields. The resulting publications were analysed based on the selection criteria, and those that met the criteria were shortlisted. A further filtering process using quality assessment criteria was applied to ensure that only high-quality publications were selected as primary studies. The data required to answer the research questions were extracted and synthesised during this step.

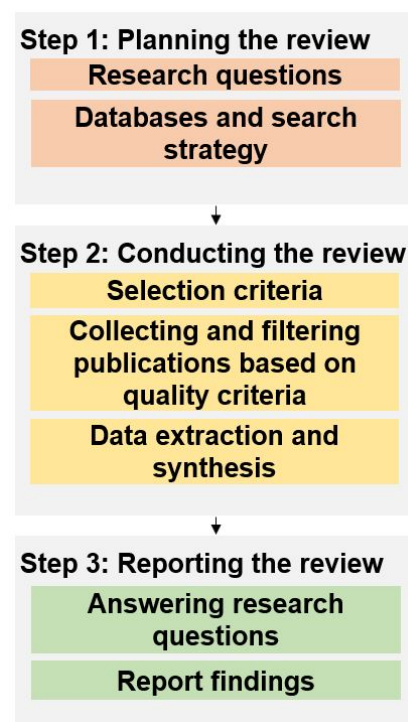

**Figure 1:** The steps involved in the process of SLR.

The third and final step was reporting the findings. The answers to research questions and the results were presented as supporting figures and tables.

### 2.2. Research Questions

The following six research questions (RQs) were formulated for this study. These questions mainly focused on exploring, collecting, and presenting recent advances in applying DL-based methodologies primarily for behaviour recognition in precision livestock farming, also called livestock behaviour recognition.

- **RQ.1:** What is the significance of behaviour recognition in livestock, and what types of problems are solved using DL for behaviour recognition?

- **RQ.2:** What approaches are used for data collection, and what is the type, quantity, and quality of data?

- **RQ.3:** What DL models and networks are used for livestock behaviour recognition?

- **RQ.4:** What performance metrics and methodologies are used to assess the outcomes of DL models?

- **RQ.5:** Which DL models and networks were most effective for a particular problem?

- **RQ.6:** What are the challenges associated with the application of DL for livestock behaviour recognition?

## 2.3. Databases and search strategy

The following seven most popular databases were selected for this study: Google Scholar, ScienceDirect, IEEE Xplore, Scopus, Web of Science, SpringerLink, and Wiley. The initial search started by combining the keywords as search strings. At the start, the keywords: *"behaviour recognition"* AND *"deep learning"* were used to find the related studies. The results contained studies related to behaviour recognition, including humans. The search string was modified to narrow down the studies only associated with the animals by adding keywords such as *"livestock"*, *"herd"*, and *"animal"*.

Furthermore, synonymous terms like *"action"* and *"activity"* related to the keyword *"behaviour"*, *"artificial intelligence"* related to *"deep learning"*, *"analysis"*, *"detection"*, and *"classification"* related to *"recognition"* were found in the literature and added to improve the search strings that resulted in the following general search string: *("action" OR "activity" OR "behaviour") AND ("artificial Intelligence" OR "deep learning") AND ("animal" OR "herd" OR "livestock") AND ("analysis" OR "classification" OR "detection" OR "recognition")*.

Subsequently, the resulting general string had to be modified based on each database. ScienceDirect only allows a maximum of eight Boolean characters, and Google Scholar has a specific limit for the maximum number of characters used in a search. Thus, the keywords were adjusted accordingly. The search strings were used in each database's abstract, article title, and keywords fields. However, Wiley and Springer Link do not allow searching in the abstract, article title, and keywords fields. Thus, the search string was used to find keywords anywhere in the publications. Because of this, a high number of unrelated publications were initially selected (as shown in Figure 2), which were then filtered further using selection and quality criteria. Overall, the initial search through the chosen databases yielded 1101 publications. The search strings used for searching publications for each of the selected databases are given below:

- Google Scholar: [("behaviour recognition" AND "livestock")] AND [("deep learning" OR "artificial intelligence") AND ("behaviour detection" OR "behaviour recognition" OR "livestock health")]

- ScienceDirect: ("deep learning" OR "artificial intelligence") AND ("livestock" OR "herd") AND ("behaviour" OR "action") AND ("recognition" OR "analysis" OR "classification")

- IEEE Xplore: ("deep learning" OR "artificial intelligence") AND ("livestock" OR "herd" OR cattle) AND ("behaviour" OR "action" OR health) AND ("recognition" OR "analysis" OR "detection" OR "classification")

- Scopus: TITLE-ABS-KEY (("deep learning" OR "artificial intelligence") AND ("livestock" OR "herd" OR cattle) AND ("behaviour" OR "action" OR health) AND ("recognition" OR "analysis" OR "detection" OR "classification")) AND (LIMIT-TO (DOCTYPE, "ar"))

- Web of Science: TI=(("deep learning" OR "artificial intelligence") AND ("livestock" OR "herd" OR cattle ) AND ("behaviour" OR "action" OR health) AND ("recognition" OR "analysis" OR "detection" OR "classification")) OR AB=(("deep learning" OR "artificial intelligence") AND ("livestock" OR "herd" OR cattle ) AND ("behaviour" OR "action" OR health) AND ("recognition" OR "analysis" OR "detection" OR "classification")) OR AK=(("deep learning" OR "artificial intelligence") AND ("livestock" OR "herd" OR cattle ) AND ("behaviour" OR "action" OR health) AND ("recognition" OR "analysis" OR "detection" OR "classification"))

- SpringerLink, Wiley: ("deep learning" OR "artificial intelligence") AND ("livestock" OR "herd" OR cattle) AND ("behaviour" OR "action" OR health) AND ("recognition" OR "analysis" OR "detection" OR "classification") (anywhere)

## 2.4. Selection Criteria

The initial search resulted in the collection of many irrelevant publications. Therefore, inclusion and exclusion criteria were defined to select publications containing information related to the research questions. Each publication was judged against the selection criteria. A publication was selected if all the exclusion criteria were false and the inclusion criteria were true (Kitchenham and Charters, 2007). Consensus against each publication was developed using the Cohen Kappa statistic (Cohen, 1968). A total of 126 publications were selected out of 1101 initial publications. The exclusion criteria used in this study are given below:

1. The publication is not related to deep learning for livestock behaviour recognition.
2. The publication is either duplicated or retrieved from another database.
3. The publication is not written in English, and the full text of the study is not available.
4. The publication is a book chapter, conference abstracts, data articles, mini-reviews, short communications, thesis, review, or survey article.
5. The publication is a pre-print or not peer-reviewed.
6. The publication was published before 2012.

The inclusion criteria used in this study are given below:

1. The publication is related to the application of DL-based methodologies for behaviour recognition in livestock.
2. The publication is a primary study.

## 2.5. Collecting and filtering publications

The 126 publications selected after the selection criteria were further accessed to ensure the selection of high-quality primary studies. For this, quality criteria based on assessment questions were chosen from the study (Kitchenham et al., 2009). Each publication was given a score of 1(yes), 0 (no), or (0.5) (partial) against each question of the quality criteria. The total score for each publication was calculated, and publication scoring less than three was excluded. A total of 44 publications were selected as primary studies as a result of the application of quality criteria. The assessment questions of the quality criteria used in this study were:

1. Are the aims & objectives of the study clearly stated?
2. Are the study's scope, methodology, and experimental design clearly defined?
3. Is the research process and methodology documented appropriately?
4. Are all study questions answered?
5. Are negative findings presented?
6. Do the conclusions correspond to the study's goals and purpose?

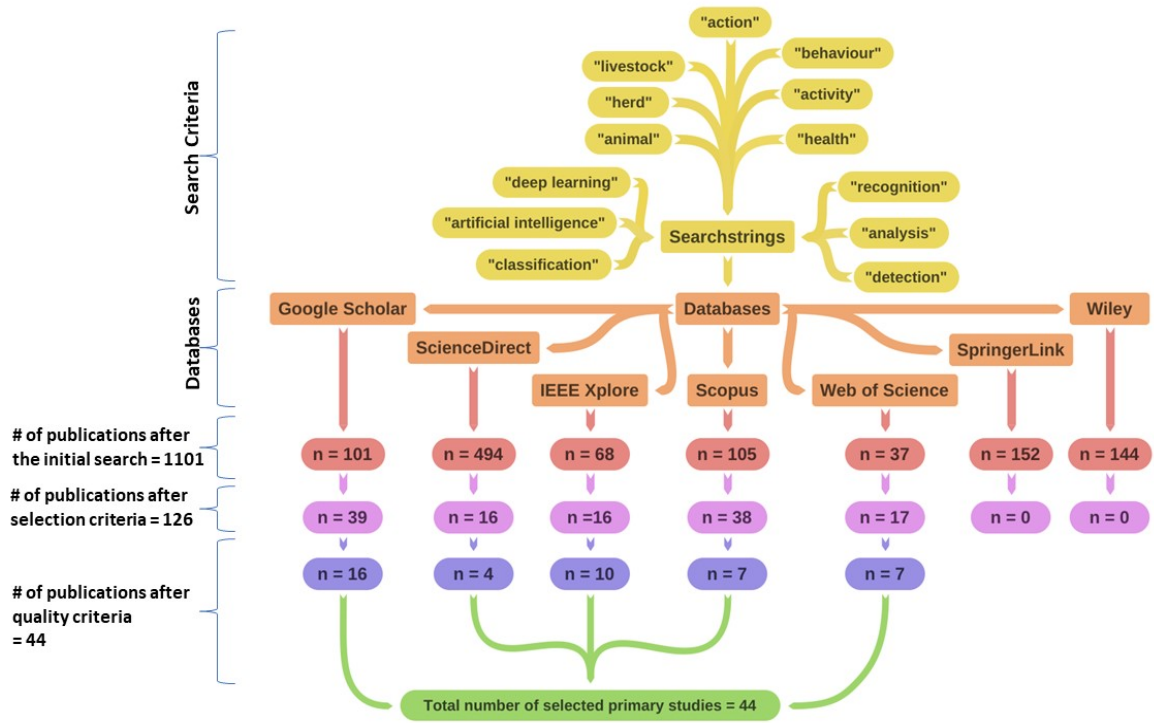

**Figure 2:** The process for the selection of primary studies.

Figure 2. shows the overall process for the selection of primary studies.

### 2.6. Data extraction and synthesis

The 44 selected primary studies were gathered and thoroughly researched to extract the relevant data related to each research question. Table 2 presents the details of the selected primary studies. A spreadsheet was used to list all the primary studies in rows against each research question in columns. Data were extracted and summarised. The extracted data focused on answering the research questions, including objectives, type of behaviour recognition problems, global research trend for livestock behaviour recognition, ethogram defining each behaviour, data collection and the type, quantity, and quality of data, DL models and networks, performance evaluation metrics, the details regarding the year and journal of publications, and challenges associated with the application of DL for livestock behaviour recognition. Finally, the extracted data were synthesised to answer each research question. The results of this SLR study are presented in the next section.

## 3. Results
### 3.1. Significance of behaviour recognition in livestock and types of problems (RQ. 1)

In recent years the task of action recognition has gained significant attention in computer vision. So far, the study of action recognition has been limited to humans (Kong and Fu, 2022). While notable progress has been made, the action recognition task is still a complex problem. With technological advancements and AI paving the way to almost every field of science, the concept of action recognition has been recently adopted to study animal behaviours. The study of animal behaviours helps to understand how a particular animal behaves under diverse circumstances. Researchers have discovered an important link between changes in animal behaviour and their health and welfare. Keeping track of animal behaviour has become an essential element of health monitoring in intensive farming systems. Although there is a great deal of interest in using action recognition to differentiate animal behaviour, it is crucial to understand the difference between an action and behaviour.

**Table 2**
Details of the selected primary studies.

| No. | Source | Article Title | Reference |
|---|---|---|---|
| 1 | Google Scholar | Mounting behaviour recognition for pigs based on deep learning | (Li et al., 2019) |
| 2 | | Pig mounting behaviour recognition based on video spatial–temporal features | (Yang et al., 2021) |
| 3 | | Automatic recognition of sow nursing behaviour using deep learning-based segmentation and spatial and temporal features | (Yang et al., 2018) |
| 4 | | Using an EfficientNet-LSTM for the recognition of single Cow's motion behaviours in a complicated environment | (Yin et al., 2020) |
| 5 | | Automatic recognition of feeding and foraging behaviour in pigs using deep learning | (Alameer et al., 2020b) |
| 6 | | An automatic recognition framework for sow daily behaviours based on motion and image analyses | (Yang et al., 2020) |
| 7 | | Recognition of feeding behaviour of pigs and determination of feeding time of each pig by a video-based deep learning method | (Chen et al., 2020a) |
| 8 | | Deep learning based classification of sheep behaviour from accelerometer data with imbalance | (Turner et al., 2022) |
| 9 | | A computer vision-based method for spatial-temporal action recognition of tail-biting behaviour in group-housed pigs | (Liu et al., 2020) |
| 10 | | Automatic Sheep Behaviour Analysis Using Mask R-CNN | (Xu et al., 2021) |
| 11 | | Automated detection and analysis of piglet suckling behaviour using high-accuracy amodal instance segmentation | (Gan et al., 2022a) |
| 12 | | Application of deep learning in sheep behaviors recognition and influence analysis of training data characteristics on the recognition effect | (Cheng et al., 2022) |
| 13 | | Automatic recognition of lactating sow postures by refined two-stream RGB-D faster R-CNN | (Zhu et al., 2020) |
| 14 | | Identification and Analysis of Emergency Behavior of Cage-Reared Laying Ducks Based on YoloV5 | (Gu et al., 2022) |
| 15 | | Reserve sow pose recognition based on improved YOLOv4 | (Lu et al., 2022) |
| 16 | | Posture Detection of Individual Pigs Based on Lightweight Convolution Neural Networks and Efficient Channel-Wise Attention | (Luo et al., 2021) |
| 17 | Science Direct | Cattle behavior recognition based on feature fusion under a dual attention mechanism | (Shang et al., 2022) |
| 18 | | Horse foraging behavior detection using sound recognition techniques and artificial intelligence | (Nunes et al., 2021) |
| 19 | | Deep learning-based hierarchical cattle behavior recognition with spatio-temporal information | (Fuentes et al., 2020) |
| 20 | | Activity detection of suckling piglets based on motion area analysis using frame differences in combination with convolution neural network | (Ding et al., 2022) |
| 21 | IEEE Xplore | Automatic Detection of Mounting Behavior in Cattle using Semantic Segmentation and Classification | (Noe et al., 2021) |
| 22 | | A Deep Learning-based solution to Cattle Region Extraction for Lameness Detection | (Noe et al., 2022) |
| 23 | | Video-based cattle identification and action recognition | (Nguyen et al., 2021) |
| 24 | | Data Augmentation for Inertial Sensor Data in CNNs for Cattle Behavior Classification | (Li et al., 2021) |
| 25 | | An AI-based System for Monitoring Behavior and Growth of Pigs | (Chen et al., 2020b) |
| 26 | | Toward Building a Data-Driven System For Detecting Mounting Actions of Black Beef Cattle | (Kawano et al., 2021) |
| 27 | | A comparison of autoencoder and statistical features for cattle behaviour classification | (Rahman et al., 2016) |
| 28 | | Lameness Detection in Cows Using Hierarchical Deep Learning and Synchrosqueezed Wavelet Transform | (Jarchi et al., 2021) |
| 29 | | On the Benefits of Deep Convolutional Neural Networks on Animal Activity Recognition | (Bocaj et al., 2020) |
| 30 | | Individual identification model and method for estimating social rank among herd of dairy cows using YOLOv5 | (Uchino and Ohwada, 2021) |
| 31 | Scopus | Spatiotemporal graph convolutional network for automated detection and analysis of social behaviours among pre-weaning piglets | (Gan et al., 2022b) |
| 32 | | Automatic Detection Method of Dairy Cow Feeding Behaviour Based on YOLO Improved Model and Edge Computing | (Yu et al., 2022) |
| 33 | | Basic motion behavior recognition of single dairy cow based on improved Rexnet 3D network | (Ma et al., 2022) |
| 34 | | Identification and classification for sheep foraging behavior based on acoustic signal and deep learning | (Wang et al., 2021) |
| 35 | | Using a CNN-LSTM for basic behaviors detection of a single dairy cow in a complex environment | (Wu et al., 2021) |
| 36 | | Image analysis for individual identification and feeding behaviour monitoring of dairy cows based on Convolutional Neural Networks (CNN) | (Achour et al., 2020) |
| 37 | | Deep learning and machine vision approaches for posture detection of individual pigs | (Nasirahmadi et al., 2019) |
| 38 | Web of Science | C3D-ConvLSTM based cow behaviour classification using video data for precision livestock farming | (Qiao et al., 2022) |
| 39 | | Automatic behavior recognition of group-housed goats using deep learning | (Jiang et al., 2020) |
| 40 | | Deep learning-based cattle behaviour recognition using joint time-frequency data representation | (Hosseininoorbin et al., 2021) |
| 41 | | Automated recognition of postures and drinking behaviour for the detection of compromised health in pigs | (Alameer et al., 2020a) |
| 42 | | Pecking activity detection in group-housed turkeys using acoustic data and a deep learning technique | (Nasirahmadi et al., 2020) |
| 43 | | Computer Vision Applied to Detect Lethargy through Animal Motion Monitoring: A Trial on African Swine Fever in Wild Boar | (Fernández-Carrión et al., 2020) |
| 44 | | Automatic recognition of lactating sow postures from depth images by deep learning detector | (Zheng et al., 2018) |

An action is defined as the act of doing something. On the contrary, a behaviour is a collection of diverse actions, sometimes repeated ones. In other words, while an action might be instantaneous, a behaviour is a set of actions that occur over a continuous interval. As a result of the addition of a temporal dimension, recognising behaviours becomes a considerably more challenging problem than simply recognising actions.

Although the DL approaches used in human action recognition can be tailored and utilized for animal behaviour analysis, there are several important differences and concerns. For example, animal behaviour data is often multi-modal. It can contain data in the form of videos, images, and signals from different times of the day under different weather conditions. This necessitates specialized preprocessing and feature extraction approaches designed specifically for animal behaviour analysis. Furthermore, animal behaviour has traits that are specific to a certain specie requiring the creation of specialised DL models that reflect these distinct patterns. The ethical constraints and restrictions in collecting the data for animals limit the availability of sufficient labelled data for training affecting the performance and generalisation of DL models.

In 44 primary studies, researchers reported 13 behaviour recognition problems. These problems were related to the behaviours such as Feeding, Posture, Motion, Nursing, Mounting, Drinking, Pecking, Tail-biting, Foraging, Social, Emergency, Lameness, and Suckling. Each of these behaviours has a significant connection with animal health and welfare. Feeding behaviour has been found to assess impaired health and predict diseases like ketosis and mastitis. Periodic evaluation of feeding behaviour helps monitor livestock's health and production status at the individual and farm levels. Recognising abnormal posture in animals on time can assist in limiting disease transmission, reducing the usage of veterinary antibiotics, and increasing the economic advantages of commercial farms. Motion and activity-related behaviours are critical predictors of the physical health status of animals and farming conditions. Drinking behaviour is linked with milk production, control of body temperature, and adequate feed consumption. Mounting behaviour is helpful in the detection of oestrous periods, and it can help improve animal reproductive performance. Nursing behaviour is closely related to starvation and influences milk output and animal growth. It can also indicate diseases or injuries related to the udder. Tail biting is one of the most harmful behaviours affecting animal welfare and production. Early lameness identification has increased animal welfare, resulting in economic and health benefits. Social behaviours are crucial indicators of animal growth and health. Emergency behaviours such as trampling in ducks are linked with a high rate of injuries. Suckling behaviour might reflect an animal's physical health. Animals with higher activity levels have been observed to have a lower mortality rate. Furthermore, the time allotted for suckling or udder massage might indicate hunger or feed intake and can be used to anticipate animal growth.

Each of these behaviour recognition problems is divided into subsequent classes. A total of 44 classes associated with these problems were reported in the primary studies. Table

**Table 3**
Type of behaviour recognition problems and associated behaviour classes.

| Type of problem | Behaviour classes |
|---|---|
| Feeding behaviour recognition | Eating, Grazing, Chewing, Ruminating, Ruminating standing, Ruminating lying, Bite |
| Posture recognition | Resting, Lying, Lying down, Lying on the belly, Lying on the side, Lateral lying, Sternal lying, Standing, Standing up, Sitting, Sternal recumbency, Ventral recumbency, Lateral recumbency, Sleeping |
| Motion recognition | Walking, Moving, Moving tail, Moving head, Inactive, Lethargy |
| Nursing behaviour recognition | Nursing interaction |
| Mounting behaviour recognition | Mounting interaction |
| Drinking behaviour recognition | Drinking interaction |
| Pecking activity recognition | Peck or Non-peck |
| Tail-biting behaviour recognition | Tail-biter (biter), Tail-bitten (victim) |
| Foraging behaviour recognition | Non-nutritive / Non-feeding visits |
| Social behaviour recognition | Searching, Social licking, Grooming, Exploring, Social nosing, Fighting, Playing |
| Emergency behaviour recognition | Neck extension, Trample, Spreading wings |
| Lameness recognition | Lameness |
| Suckling behaviour recognition | Suckling or Non-suckling |

3 presents the type of behaviour recognition problems and associated behaviour classes reported in the primary studies.

The first and most crucial step to collecting data and properly distinguishing between distinct classes is to define each class. It is critical to establish the limits of a particular behaviour class. Ethologists organise these behavioural classes into what is known as an ethogram. In this SLR study, an ethogram summarising the definition of each behaviour class reported in the primary studies is presented in Table 4. Due to the lack of clear criteria for defining a particular class's boundaries, several definitions are connected with each behaviour class. The criteria employed by the researchers were based on their observations, practical experiences, and the surrounding farm environment. A deep learning model developed with one definition for the same class might fail to perform in an environment with a different definition. It is essential, and so far, an unaddressed research question, to establish a specified ethogram detailing behaviour classes expressly for developing deep learning models for livestock behaviour recognition, which will also facilitate the performance comparison of these models on equal merits.

### 3.1.1. Distribution of primary studies and global trend for livestock behaviour recognition

Figure 3 shows the distribution of the selected primary studies by year and journal of publication. This SLR study focused on articles published between 2012 and 2022. From 2012 to 2015, no article explicitly focused on applying deep learning to livestock behaviour recognition. Since 2016, there has been an increasing trend. This increase was because these were the years when the DL learning-based approaches started to achieve more prominent results on problems such as image and video classifications, object detection, target tracking, and regression. The works before 2016 mostly concentrated on manual feature selection and extraction methodologies that lack scalability.

The highest number of articles were published in 2022. A total of 12 journals were significant in publishing related articles, with 04 publishers. The journal with the most published articles has been Computer and Electronics in Agriculture, followed by IEEE Conference Proceedings and Biosystems Engineering.

A Sankey chart summarising the global research trend is shown in Figure 4. The chart is organised into five sections, including information on the countries, livestock types, behaviour levels, and the type of behaviour recognition problem reported in primary studies. China was found to be the research leader in the application of DL for livestock behaviour recognition, with 43.75%, followed by Australia (15.91%), Japan (10.8%), the United Kingdom (UK) (6.25%), the United States of America (USA) (5.11%), and South Korea (4.55%), making up the top five out of a total of 16 countries. The research was primarily focused on seven different types of livestock, with cattle receiving the most attention (43.18%), followed by pigs (35.23%), sheep (10.8%), goats (5.68%), horses (2.27%), turkey (1.7%), and ducks (1.14%). Certain countries have focused on a specific type of livestock, such as China focused on all reported livestock types except turkey and horses. Australia focused primarily on cattle and sheep. Japan and South Korea only on cattle. The UK on pigs and turkey, and the USA on pigs and horses. No specific correlation between animal production and the increased amount of research was found. However, as cattle, pigs, sheep, and goats are consumed mostly as sources of food and are in the majority at livestock farms, the research on these animals was more prominent than others.

Furthermore, there are different levels at which livestock behaviours can be categorised. This SLR study has categorised this into primary and secondary behaviour levels. The primary behaviour level is divided into two categories: individual and group. The individual level refers to behaviours solely related to a single animal, for example, Motion, Posture, Pecking, Emergency, and Lameness. However, the group level refers to behaviours that include multiple animals, for example, Mounting, Social, Suckling, Nursing, and Tail-biting. Some behaviours can be at individual or group levels, for example, Feeding, Drinking, and Foraging.

**Table 4**
Ethogram outlining behaviour classes found in the primary studies.

| Behaviour classes | Description | Reference |
| --- | --- | --- |
| Feeding /Eating | Standing with the head in the feed tray OR Lowering head OR Keeping head in the food area and performing eating movement OR Head inside a food trough OR Keeping head in the food area OR Biting and chewing on its food, and rooting with snout in the feeder OR Head is downward and the food is available in the feeder OR When the bounding box of any pig's head overlaps the region of crib | (Fuentes et al., 2020), (Yang et al., 2020),(Alameer et al., 2020b),(Chen et al., 2020a),(Achour et al., 2020),(Chen et al., 2020b),(Qiao et al., 2022),(Yu et al., 2022) |
| Chewing | Chewing, and head raised | (Rahman et al., 2016) |
| Ruminating | Chewing regurgitated bolus OR Mouth chews repeatedly without other vigorous movement | (Fuentes et al., 2020),(Wu et al., 2021) |
| Ruminating standing | Ruminating, and standing | (Rahman et al., 2016) |
| Ruminating lying, | Ruminating, and lying or sitting | (Rahman et al., 2016) |
| Resting | Sitting on the floor OR Resting, and lying OR Resting, and standing with head up | (Fuentes et al., 2020),(Rahman et al., 2016) |
| Lying | Abdomen clings to the ground without other strenuous movement OR The posture after kneeling | (Wu et al., 2021),(Ma et al., 2022) |
| Lying down | Action from standing to lying down | (Fuentes et al., 2020) |
| Lying on the belly | Lying with limbs are folded under the body | (Nasirahmadi et al., 2019) |
| Lying on the side | Lie in a fully recumbent position with limbs extended | (Nasirahmadi et al., 2019) |
| Lateral lying | The side of the trunk of the pig is in contact with the floor | (Alameer et al., 2020a) |
| Sternal lying | The chest / sternum of the pig is in contact with the floor | (Alameer et al., 2020a) |
| Standing | Upright body position on extended legs with hooves only in contact with the floor OR Legs are straight and supporting the body OR The head (upward) indicates that the dairy cow is in standing state OR Pig has feet (and possibly snout) in contact with the pen floor OR Legs remain vertical without other strenuous movement OR The position of body and four legs is unchanged OR The posture before kneeling on the ground | (Zheng et al., 2018),(Zhu et al., 2020),(Fuentes et al., 2020), (Achour et al., 2020), (Alameer et al., 2020a),(Wu et al., 2021),(Qiao et al., 2022),(Ma et al., 2022),(Lu et al., 2022) |
| Standing up | Action from lying down to standing | (Fuentes et al., 2020) |
| Sitting | Partly erected on stretched front legs with caudal end of body contacting the floor OR Only the feet of the front legs and the posterior portion/bottom of the pig body are in contact with the floor | (Zheng et al., 2018), (Zhu et al., 2020), (Alameer et al., 2020a), (Lu et al., 2022) |
| Sternal recumbency | Lying on abdomen/sternum with front and hind legs folded under the body; udder is totally obscured, | (Zheng et al., 2018), (Zhu et al., 2020) |
| Ventral recumbency | Lying on abdomen/sternum with front legs folded under the body and visible hind legs (right side, left side); udder is partially obscured | (Zheng et al., 2018),(Zhu et al., 2020), (Lu et al., 2022) |
| Lateral recumbency | Lying on either side with all four legs visible (right side, left side); the udder is totally visible OR Side lying flat on the ground, one shoulder on the ground, with the limbs extended laterally | (Zheng et al., 2018), (Zhu et al., 2020),(Lu et al., 2022) |
| Sleeping | Sitting on the floor and head on the ground | (Fuentes et al., 2020) |
| Walking | Moving in a standing position OR Head up and walking OR Legs move repeatedly and cow position changes greatly OR Movement for more than 3 sec | (Fuentes et al., 2020),(Wu et al., 2021),(Qiao et al., 2022),(Ma et al., 2022) |
| Moving | Moving without doing anything else | (Yang et al., 2020) |
| Moving tail | Tail movements | (Fuentes et al., 2020) |
| Moving head | Head movements | (Fuentes et al., 2020) |
| Inactive | Sitting, lying, kneeling, or standing without performing any other activity | (Yang et al., 2020) |
| Lethargy | Low values in daily motion matched with high temperature peaks | (Fernández-Carrión et al., 2020) |
| Nursing interaction | At least half of the piglets are actively manipulating the udder when the sow is lying laterally, and the period of activity exceeds 60 sec | (Yang et al., 2018), (Yang et al., 2020) |
| Mounting interaction | Two individuals mounting OR When one pig's front part was firstly placed on the other pig's body at the start of the mounting and the frame when the two pigs were completely separated was considered to be the end of the mounting OR Cow bends over another cow usually when either cow is in estrus | (Fuentes et al., 2020),(Yang et al., 2021),(Kawano et al., 2021),(Luo et al., 2021) |

Table 4: (Continue)

| Behaviour classes | Description | Reference |
| --- | --- | --- |
| Drinking interaction | Touching the drinking nipple with snout OR The pig snout is in contact with a nipple drinker, OR Stand by the water tank with its mouth in the tank | (Yang et al., 2020), (Alameer et al., 2020a),(Wu et al., 2021) |
| Peck or Non-peck | When birds struck the metallic ball (pecking object) with their beak | (Nasirahmadi et al., 2020) |
| Tail-biter | Biting a penmate's tail, with a sudden reaction of the penmate | (Turner et al., 2022) |
| Tail-bitten | Penmate is biting the subject's tail and elicits a reaction | (Turner et al., 2022) |
| Non-nutritive / Non-feeding visits | When a pig enters the feeding area with two feet without ever consuming any food OR When the cow is in the feeding zone, lifts her head away from the feed area to chew, and the next action is either to continue feeding or to leave the feeding zone | (Alameer et al., 2020b),(Yu et al., 2022) |
| Searching | Head down and walking | (Rahman et al., 2016) |
| Social licking | Licking another's body with the tongue | (Fuentes et al., 2020) |
| Grooming | Licking own body with the tongue OR Head is turned towards abdomen groom the body with the tongue | (Fuentes et al., 2020),(Qiao et al., 2022) |
| Exploring | Head is in close proximity of or in contact with the ground | (Qiao et al., 2022) |
| Social nosing | Piglet touching or sniffing any part of the head or nose of another piglet | (Gan et al., 2022b) |
| Fighting | Two or more individuals fighting OR Forceful fighting, pushing with the head, or violently biting littermates | (Fuentes et al., 2020),(Gan et al., 2022b) |
| Playing | Nudging or pushing, playing, and fighting | (Gan et al., 2022b) |
| Neck extension | The laying duck stretches its neck out of the cage from the back of the cage | (Gu et al., 2022) |
| Trample | At least one of the feet of one duck is trampling on the body of the other duck | (Gu et al., 2022) |
| Spreading wings | Duck wings spread from a certain angle to fully unfolded | (Gu et al., 2022) |
| Lameness | Estimate gait frequency, duration of gait and non-gait periods, and the lameness OR Speed of walking, arching their backs and drop their heads during walking | (Jarchi et al., 2021),(Noe et al., 2022) |
| Suckling or Non-suckling | Either mouth on teat or nose contact to udder with vertical and rhythmic head movements, consisting of udder pre-massage, milk intake/outflow, and post-massage OR Cow feeding a calf | (Gan et al., 2022a),(Fuentes et al., 2020) |

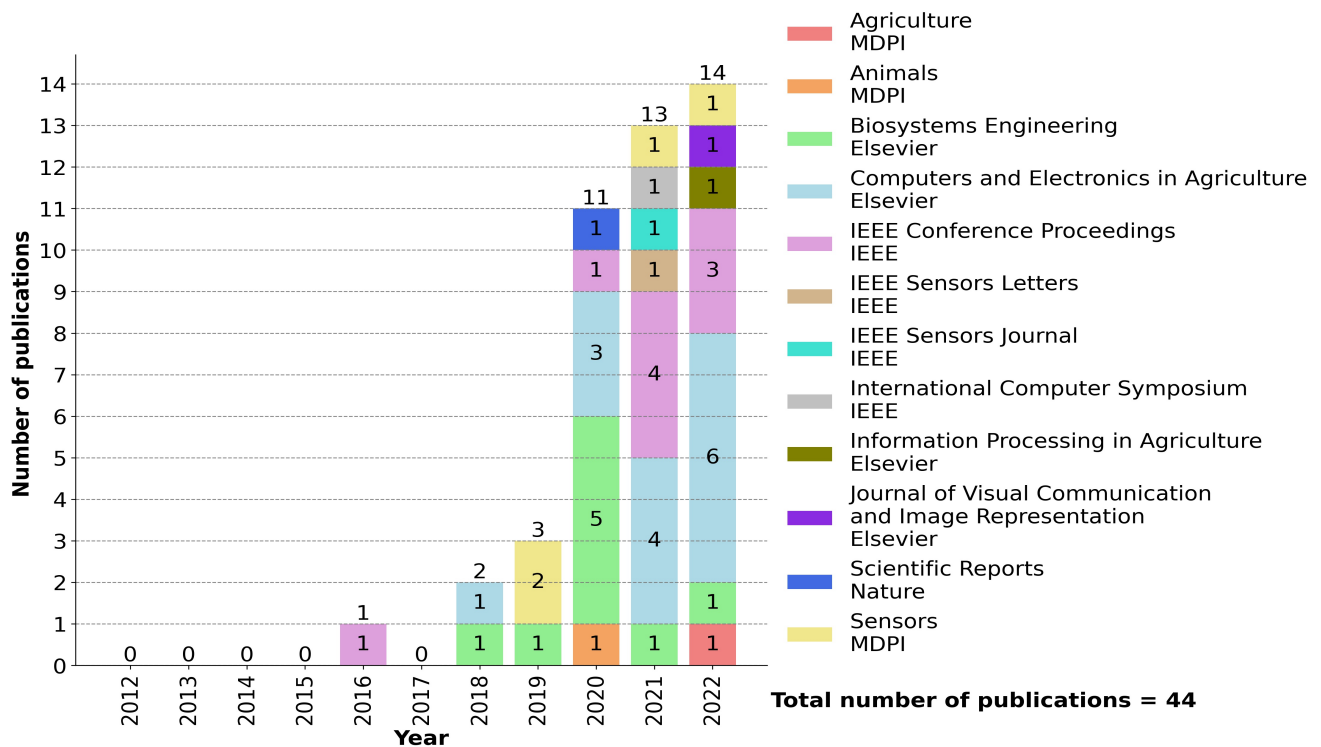

**Figure 3:** The distribution of the selected primary studies by year and journal of publication.

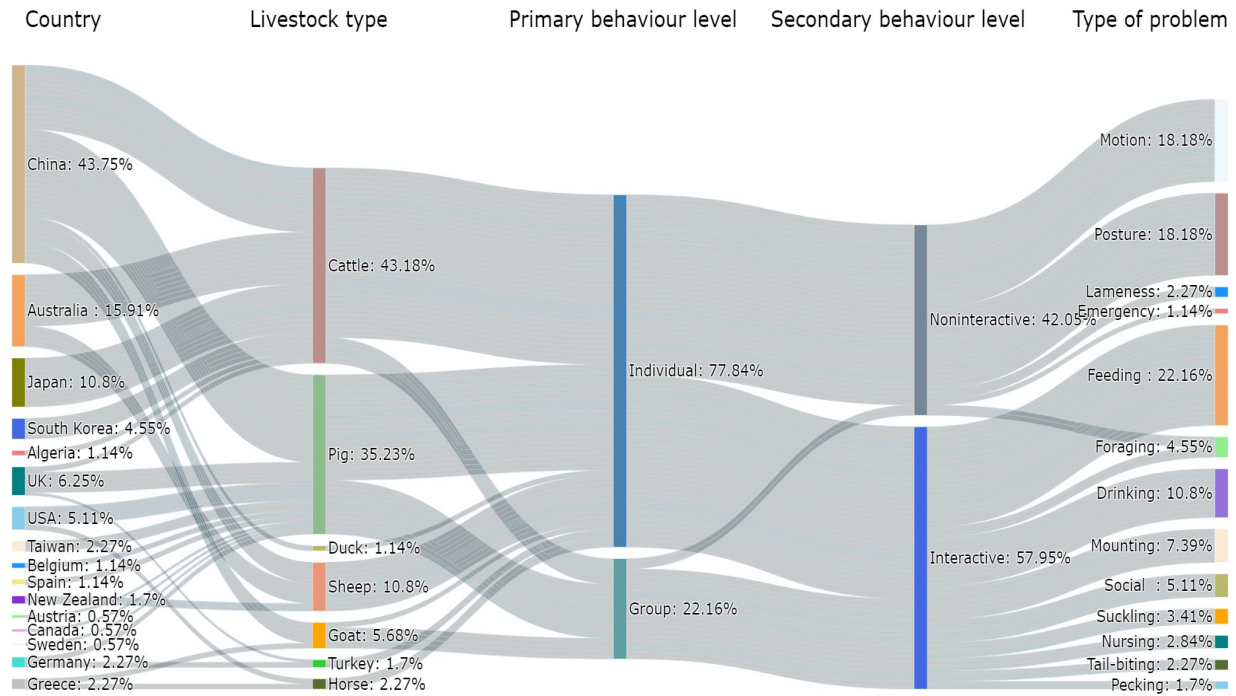

**Figure 4:** Global research trend for livestock behaviour recognition.

On the contrary, the secondary behaviour level is divided into two categories: interactive and non-interactive. The interactive level refers to behaviour where single or multiple animals interact with one another or an object from the surrounding environment, for example, Pecking, Mounting, Social, Suckling, Nursing, Tail-biting, Feeding, Drinking, and Foraging. However, the non-interactive level refers to behaviours that do not involve animals engaging with one another, for example, Motion, Posture, Emergency, and Lameness. The researchers focused more on the individual (77.84%) than group (22.16%) behaviours at the primary level and more on interactive (57.95%) than non-interactive (42.05%) behaviours at the secondary level. The majority of studies focused on individual than group behaviours. This correlates with the fact that in computer vision, individual-level object detection is easier than grouped-level object detection because of the challenges related to spatial localization, occlusion and adhesion. Group behaviours were found to be complex and hard to recognize. Similarly, the individual non-interactive behaviours were studied more than the group interactive (Figure 4).

Feeding behaviour recognition was the most prevalent of the 13 types of behaviour recognition problems reported in primary studies, with 22.16%, followed by Posture and Motion (18.18%), Drinking (10.8%), Mounting (7.39%), Social (5.11%), Foraging (4.55%), Suckling (3.41%), Nursing (2.84%), Lameness and Tail-biting (2.27%), Pecking (1.7%), and Emergency (1.14%).

Feeding behaviour was the most focused behaviour among the others since feeding was found to be directly linked with nutritional intake and the overall well-being of the animals. Furthermore, as feeding is individual behaviour, the majority of the studies focused on recognizing it than complex group behaviours such as mounting, social, and suckling.

### 3.2. Data collection and the type, quantity, and quality of data (RQ. 2)

Data is one of the most crucial elements on which the performance of a DL model is highly dependent. DL models are considered to be data-hungry. Although the DL community agrees that the more data there is, the more accurate the model will be, it is still being determined how much data is acceptable and what the limit should be. These are open research questions. Most of the time, researchers use their expertise and experience to define the parameters of a good dataset. Data scientists and AI developers dedicate about 70% of their time to analysing, pre-processing, and structuring data. The remainder is spent on model selection, training, testing, and implementation procedures. A similar pattern was found in the 44 primary studies.

There were particular reasons for choosing a specific data type. The choice of data either signals or images was solely based on the approach that was adopted by the researchers. The visionless or contact-based approaches used signals data as the source of the data were physical sensors. However, the vision-based or contactless approaches used images or videos as the source of data were the camera. Mostly, the signals were used for the behaviours such as Motion, Posture, and Lameness where recording the motion data was a crucial parameter. However, no tangible explanations were found regarding the data's quality and quantity. A different number of samples, frequency ranges, image sizes, and FPS were used by different studies even for the recognition of the same type of behaviour. Regardless of this, in this SLR study, a summary of data collection instruments, data type, data quantity, and data quality, along with the type of behaviour recognition problems solved, the reported DL models and networks, their performance, and the number of classes used are synthesised and presented in Table 5 and 6.

Generally, the approaches used in primary studies can be classified into two categories: visionless or contact-based and vision-based or contactless. In the contact-based approach, the studies reported using body-mounted sensors such as inertial measurement units (IMUs). These sensors directly contacted the body of the animals and were reported to affect their health and welfare over time. In addition, these sensors were prone to read false readings, such as noise, due to the surrounding environment. For this approach, the data was recorded generally as accelerometer signals, with the lowest samples at 21,924 and the highest at 3,500,000, with a sampling rate ranging from 10Hz to 44KHz.

On the contrary, non-intrusive methods were developed in the contactless approach without coming in contact with the animal body using camera systems. The data was recorded in images or videos, and data quantity and quality varied across different studies. The lowest number of images recorded was 108, and the highest was 630,000. The image quality was reported to be 224 x 224 pixels at a minimum and 3840 x 2160 pixels at a maximum, with a frame rate ranging from 01 FPS to 30 FPS.

### 3.3. DL models and networks for livestock behaviour recognition (RQ. 3)

In the primary studies, a total of 23 various DL models were reported for addressing the 13 different types of behaviour recognition problems. These models include different variants of YOLO, CNN, R-CNN, DNN, LSTM, and some combinations of models such as CNN-LSTM and C3D-ConvLSTM. Figure 5 shows the DL models reported in the primary studies and the number of times they were used for each behaviour recognition problem. Among 23 DL models and their variants, the most popular were CNN, Faster R-CNN, YOLOv5, YOLOv4, and CNN-LSTM. Although selecting the DL model for a specific type of problem is challenging with no defined rules, initial research found a link between model selection and the type of behaviour recognition problem. YOLOv5 was primarily used for drinking behaviour recognition since drinking behaviour is commonly regarded as an individual and interactive level object detection problem . It has provided superior performance for a wide range of applications due to its incredible efficiency and power in detecting objects across various domains.

Similarly, CNN was used for Feeding, Foraging, and Motion recognition since they are highly effective in analyzing videos or images by identifying unique feeding, foraging, or motion patterns based on spatial information within the data. However, LSTM was used for Lameness recognition since the available data was primarily in time series and was acquired using physical IMU sensors. Faster R-CNN was mainly used for Mounting and Posture behaviour recognition since it uses an attention mechanism based on RPN to extract regions of interest from images, allowing it to identify behaviours at the individual and group levels, whether interactive or non-interactive. Faster R-CNN, on the contrary, is computationally demanding when it comes to real-time implementation.

As stated previously, behaviour is defined across a continuous period. As a result, the time domain becomes an essential aspect in recognising behaviours. Although the state-of-the-art is currently confined to instantaneous frame-by-frame recognition, some models consider the time factor. C3D-ConvLSTM is one of these models used in studies to recognise Feeding, Motion, and Social behaviour. The C3D-ConvLSTM is the LSTM with additional 3D convolutional layers. A DL model is often employed in two dimensions; however, in three dimensions, an additional dimension is introduced to increase the model's accuracy by expanding the convolution along with the temporal dimensions, allowing it to learn both discriminative visual features and temporal correlations. Another example is the TSN, which uses segment-based sampling to ensure that the sampled video clips are distributed equally over the temporal dimension, providing another critical indication for action understanding. TSN was reported in one study for Drinking behaviour recognition only.

Figure 6 shows the DL networks reported in the primary studies and the number of times they were used for each behaviour recognition problem. A total of 24 various DL networks were reported in the primary studies. Most of the DL networks were custom-built, with researchers designing their network architecture rather than using pre-defined ones. Furthermore, the top five pre-defined network architectures were VGG16, CSPDarknet53, GoogLeNet, ResNet101, and ResNet50. Figure 7 shows the combinations of DL models and networks reported in the primary studies. YOLOv5 was used mostly with CSPDarknet53 as the backbone network, followed by CNN with custom-built networks, and Faster R-CNN with ResNet50.

### 3.4. Performance metrics and methodologies (RQ. 4)

Researchers used ten different performance metrics to evaluate the effectiveness of DL models. Figure 8 shows the performance metrics and the number of times they were reported in the primary studies. Precision and Accuracy were the most used performance metrics, followed by the F1-score, mean Average Precision (mAP), Recall, Sensitivity, Specificity, Average Precision (AP), Intersection over Union

**Table 5**
Summary of data collection, type, quantity, and quality.

| Reference | Type of problem | Data collection | Data type | Data quantity | Data quality |
|---|---|---|---|---|---|
| (Rahman et al., 2016) | Feeding, Posture, Motion | Neck mounted collar with sensors, Honeywell HMC6343 | Accelerometer signals | 8,62,500 samples | 10Hz sampling rate |
| (Zheng et al., 2018) | Posture | Microsoft Kinect v2 sensor | RGB & depth images | 356,000 images | 512 x 424 px |
| (Yang et al., 2018) | Nursing | Hikvision camera (DS-2CD1321D-I) | Videos, RGB images | 421,972 images | 1920 x 1080 px, 10 FPS |
| (Nasirahmadi et al., 2019) | Posture | Two top view cameras (VIVOTEK IB836BA-HF3, Hikvision DS-2CD2142FWD-I) | RGB Images | 4,900 images | 1280 x 720 px |
| (Li et al., 2019) | Mounting | GigE camera (Allied Vision Technologies, Manta G-282C, Nürnberg, Germany) | Videos, RGB images | 1,500 images | 1936 x 1458 px, 2 FPS |
| (Zhu et al., 2020) | Posture | Microsoft Kinect v2 sensor | RGB & depth images | 18,133 pairs of images | 1080 x 1920 px, 5 FPS |
| (Yang et al., 2020) | Feeding, Drinking, Nursing, Motion | Hikvision camera (DS-2CD1321D-I) | Videos, RGB images | 630,000 images | 960 x 540 px, 5 FPS |
| (Nasirahmadi et al., 2020) | Pecking | Microphone (Monacor VB-120MIC) Camera (TosiNet Realtime 2K 4 MPPoE-IP-camera) | Audio signals, Videos | 13,100 sound clips | 44,100 Hz sampling rate |
| (Turner et al., 2022) | Tail-biting | 6 IP camera (GV-BX 1300KV, Geovision Inc., Taipei, Taiwan) | Videos, RGB images | 8 hr video (247 tail biting events lasting 1s to 10s) | 1280 x 720 px, 30 FPS |
| (Alameer et al., 2020b) | Feeding, Foraging | Two cameras (Microsoft Kinect for Xbox One, Microsoft, Redmond, Washington, USA) | RGB & grayscale images | 42,778 images | 640 x 360 px, 25 FPS |
| (Chen et al., 2020a) | Feeding | IR outdoor dome security camera (CTP-TLVA29AV, Cantek Plus, USA) | Videos | 44 hr, 1 sec clips for each class | 1180 x 830 px, 30 FPS |
| (Achour et al., 2020) | Feeding | Raspberry Pi 3 Model B connected to a USB webcam (Yudanny USB HD Webcam) | Videos, RGB images | 19 hrs, 25,352 images (4 datasets) | 480 x 640 px |
| (Alameer et al., 2020a) | Posture, Drinking | Camera (Microsoft Kinect for Xbox One, Microsoft, Redmond, Washington, USA) | Videos, RGB images | 11,3379 images | 640 x 360 px, 25 FPS |
| (Fuentes et al., 2020) | Posture, Feeding, Social, Mounting | Cameras | Videos, RGB & opitcal flow images | 350 videos, 12 min each | 1920 x 1080 px, 1 FPS |
| (Yin et al., 2020) | Motion | Cameras (SONY HDR-CX290 & a network monitoring camera system YW7100HR09-SC62-TA12) | Videos, RGB images | 1,009 videos, 90 sec each, 2,270,250 images | 512 x 512 px, 25 FPS |
| (Bocaj et al., 2020) | Feeding, Motion | Public datasets | Accelerometer signals | Two datasets 87621 samples, 6 subjects (Horse) 86557 samples, 5 subjects (Goat) | 12 Hz sampling rate 100 Hz sampling rate |
| (Fernández-Carrión et al., 2020) | Motion | Fixed dome cameras | Videos, RGB images | 1,000 images | 640 x 360 px, 6 FPS |
| (Jarchi et al., 2021) | Lameness | Wearable device based on the Intel QuarkSE microcontroller C1000 integrating Bosch BMI160 (Bosch-sensortec.com, 2016) | Accelerometer signals | 2,04,999 samples | 16 Hz sampling rate |

| Reference | Type of problem | Data collection | Data type | Data quantity | Data quality |
|---|---|---|---|---|---|
| (Wu et al., 2021) | Posture, Drinking, Motion | Hikvision DS-2DM1-714 dome webcam (Hikvision Digital Technology Co. Ltd., Hangzhou, China) | Videos, RGB images | 4566 videos, 63 hr, 10-55 sec each | 224 x 224 px, 25 FPS |
| (Chen et al., 2020b) | Feeding, Drinking | Wide angle fish eye camera | Videos, RGB & grayscale images | 1 day | 2304 x 1296 px |
| (Nunes et al., 2021) | Foraging | LK-SC100B micro-camera (LKSUMPT, Shenzhen, China) | Audio, Video | 20 min, 2309 events, 0.3 to 0.5 sec each | 16 kHz, 32 bits |
| (Noe et al., 2021) | Mounting | Cameras | Videos, RGB images | 15 videos, 2,000 Images | 1024 x 768 px |
| (Yang et al., 2021) | Mounting | Haikang infrared network camera (DS-2CD3345-I, Hikvision, Hangzhou, China) | Videos, RGB images | 1,000 images | 2560 x 1440 px, 25 FPS |
| (Kawano et al., 2021) | Mounting | Cameras | Videos, RGB images | 12 hrs, 5,020 images | 224 x 224 px |
| (Hosseininoorbin et al., 2021) | Feeeding, Motion, Drinking | Collar tags, IMU sensor | Accelerometer signals | 19 hr, 3,500,000 samples | 50 Hz sampling rate |
| (Wang et al., 2021) | Foraging | Collar with microphone 9750 byTwin-star of Korea | Audio, Video | 21,924 samples | 44,100 Hz sampling rate |
| (Li et al., 2021) | Feeding, Motion | Public dataset | Accelerometer signals | 5,30,485 samples | 25 Hz sampling rate |
| (Nguyen et al., 2021) | Drinking | Three cameras (GoPro 5 Black) | Videos, RGB images | 1,715 videos, 64 hr | 256 x 256 px |
| (Luo et al., 2021) | Posture, Mounting | IR camera | Videos, RGB images | 22,509 images | 2304 x 1296 px, 15 FPS |
| (Xu et al., 2021) | Posture | 5 turret cameras (MR832, 1080p) | Videos, RGB images | 108 images | 1280 x 720 px, 15 FPS |
| (Qiao et al., 2022) | Feeding, Social, Motion | DS-2DM1-714 integrated IP camera (Hikvision Inc., Hangzhou, China) | Videos, RGB images | 40 hr, 484( calf) & 236(cow) videos | 704 × 576 px, 25 FPS |
| (Ding et al., 2022) | Suckling | Camera(DS-2CD3346WD-I, HIKVISION, Hangzhou, China) | Videos, RGB images | 5,000 images | 640 x 640 px, 24 FPS |
| (Ma et al., 2022) | Motion | Cameras (Shenzhen, Yiwei Ruichang Technology Co., Ltd.,YW7100HR09-SC62- TA12 | Videos, RGB images | 406 videos, 15-30 sec each, 2,56,500 images | 224 x 224 px, 25 FPS |
| (Gu et al., 2022) | Emergency | Camera | RGB images | 5,560 images | 640 x 640 px |
| (Gan et al., 2022b) | Social | Camera (DS-2CD1321D-I, Hikvision, Hangzhou, China) | Videos | 100 videos, 30 sec each | 960 x 540 px, 5 FPS |
| (Noe et al., 2022) | Lameness | Camera | RGB images | 2,000 images | 512 x 512 px |
| (Turner et al., 2022) | Posture, Motion, Feeding | Jaw mounted ActiGraph sensors (ActiGraph, Pensacola, Florida,USA) & ear mounted activity sensors (Axivity Ltd, Newcastle,UK) | Accelerometer signals | Two datasets 29,179 & 2,54,087 samples | 30 & 25 Hz sampling rate |
| (Shang et al., 2022) | Posture, Motion, Feeding | Pucblic dataset, Camera recorded custom dataset | Videos, RGB images | 61 videos, 4,360 images | 500 x 500 px |
| (Yu et al., 2022) | Feeding | Two ZED2 depth camera (STEREOLABS) | Videos,RGB images | 80 videos, 10,288 images | 1280 x 720 px, 30 FPS |
| (Cheng et al., 2022) | Posture, Drinking, Feeding | Cameras (HIKVISION, Hangzhou, China) | Videos,RGB images | 12 days, 100 hr | 1920 x 1080, 12 FPS |
| (Lu et al., 2022) | Posture | Camera | Videos, RGB images | 2,310 images | 416 x 416 px, 15 FPS |
| (Gan et al., 2022a) | Suckling | Cameras (IPX DDK-1700D, USA) | Videos | 100 videos, 60 sec each & 8 hr test video | 1024 x 768, 7 FPS |
| (Uchino and Ohwada, 2021) | Foraging, Motion, Drinking | 4 Cameras | Videos,RGB images | 3 days, 1,093 images | 3840 x 2160 px |
| (Jiang et al., 2020) | Posture, Drinking, Feeding, Motion | EZVIZ camera (HIKVISION,Hangzhou, China) | Videos, RGB images | 2,000 images | 1280 x 720 px, 15 FPS |

**Table 6**
Summary of DL models and networks, number of classes, and their performance

| Reference | Type of problem | Deep Learning models | Networks | Candidate objects | Classes | Performance |
|---|---|---|---|---|---|---|
| (Rahman et al., 2016) | Feeding, Posture, Motion | SAE | Custom-built | N/A | 9 | Average precision: 62% |
| (Zheng et al., 2018) | Posture | Faster R-CNN | ZFNet | RPN | 5 | Accuracy: 93.58% |
| (Yang et al., 2018) | Nursing | FCN | VGG16 | N/A | 2 | Accuracy: 96.8% |
| (Nasirahmadi et al., 2019) | Posture | Faster R-CNN, R-FCN, SSD | Inception V2, ResNet50, ResNet101 | N/A | 3 | mAP: 91% using Faster R-CNN & Inception V2 |
| (Li et al., 2019) | Mounting | Mask R-CNN | ResNet50, ResNet101 | RPN | 2 | Accuracy: 91.47% |
| (Zhu et al., 2020) | Posture | Faster R-CNN | ZFNet | RPN | 5 | Average precision: 95.47% using concatenation fusion method |
| (Yang et al., 2020) | Feeding, Drinking, Nursing, Motion | FCN | AlexNet, VGG16, GoogLeNet | N/A | 6 | Accuracy: 97.49% (drinking), 95.36% (feeding), & 88.09% (nursing) |
| (Nasirahmadi et al., 2020) | Pecking | CNN | Custom-built | N/A | 2 | Accuracy: 96.8% |
| (Turner et al., 2022) | Tail-biting | SSD, CNN-LSTM | VGG16, ResNet-50 | N/A | 2 | Accuracy: 96.35% using ResNet-50 |
| (Alameer et al., 2020b) | Feeding, Foraging | CNN | GoogLeNet, Sc-GoogLeNet | N/A | 7 | Accuracy: 99.4% |
| (Chen et al., 2020a) | Feeding | CNN-LSTM | Xception | N/A | 2 | Accuracy: 98.4% |
| (Achour et al., 2020) | Feeding | 4 CNN models | Xception | N/A | 2, 2, 6, 17 | Accuracy: 92.61% |
| (Alameer et al., 2020a) | Posture, Drinking | Faster R-CNN, YOLO | ResNet-50 | RPN | 5 | mAP: 98.9% |
| (Fuentes et al., 2020) | Posture, Feeding, Social, Mounting | Faster R-CNN, YOLOv3 | VGG16, Darknet53 | N/A | 15 | mAp: 85.6% using saptio-temporal analysis (YOLOv3) |
| (Yin et al., 2020) | Motion | CNN-LSTM | EfficientNet, VGG16, ResNet50, DenseNet169 | BiFPN | 5 | Accuracy: 97.87% |
| (Bocaj et al., 2020) | Feeding, Motion | CNN | Custom-built | N/A | 6, 5 | Accuracy: 97.42% |
| (Fernández-Carrión et al., 2020) | Motion | CNN | AlexNet | N/A | 2 | Accuracy: 97.2% |
| (Jarchi et al., 2021) | Lameness | LSTM | Custom-built | N/A | 4 | Accuracy: 96.73% |
| (Wu et al., 2021) | Posture, Drinking, Motion | CNN-LSTM, Bi-LSTM | VGG16, VGG19, ResNet18, ResNet101, MobileNetV2, DenseNet201 | N/A | 5 | Accuracy; 98% |
| (Chen et al., 2020b) | Feeding, Drinking | Mask R-CNN | Custom-built | FPN | 3 | mAP: 89.1% |

(IoU) and Mathews Correlation Coefficient (MCC). Most of the studies used a combination of more than one metric.

Accuracy was used for 12 behaviours, including Drinking, Feeding, Foraging, Lameness, Motion, Mounting, Nursing, Pecking, Posture, Social, Suckling, and Tail-biting. AP was used for 03 behaviours, including Mounting, Posture, and Social. F1-score was used for ten behaviours, including Drinking, Emergency, Feeding, Foraging, Lameness, Motion, Pecking, Posture, Social, and Suckling. The mAP was used for 08 behaviours, including Drinking,

| Reference | Type of problem | Deep Learning models | Networks | Candidate objects | Classes | Performance |
|---|---|---|---|---|---|---|
| (Nunes et al., 2021) | Foraging | LSTM | Custom-built | N/A | 2 | mAP: 86.42% |
| (Noe et al., 2021) | Mounting | CNN | VGG16 | N/A | 2 | Accuracy: 98% |
| (Yang et al., 2021) | Mounting | Faster R-CNN | ResNet50 | RPN | 2 | Accuracy: 95.15% |
| (Kawano et al., 2021) | Mounting | CNN | ResNet50 | N/A | 4 | mAP: 80.02% |
| (Hosseininoorbin et al., 2021) | Feeeding, Motion, Drinking | DNN | Custom-built | N/A | 9 | F1 Score: 89.3% |
| (Wang et al., 2021) | Foraging | CNNRNN | Custom-built | N/A | 5 | Accuracy: 93.17% using RNN |
| (Li et al., 2021) | Feeding, Motion | CNN | Custom-built | N/A | 5 | mAP: 95.4% |
| (Nguyen et al., 2021) | Drinking | RCNN, TSN | Custom-built | N/A | 3 | Accuracy: 84.4% |
| (Luo et al., 2021) | Posture, Mounting | Faster R-CNN, YOLOv3, YOLOv5 | ResNet50, Darknet53, CSPDarknet53, MobileNetV3, SPD-YOLO | FPN | 5 | mAP: 92.04% (YOLOv5) |
| (Xu et al., 2021) | Posture | Mask R-CNN | Custom-built | RPN | 2 | mAP: 94% |
| (Qiao et al., 2022) | Feeding, Social, Motion | C3D-ConvLSTM | C3DLSTM, RNN, Inception V3 | N/A | 5 | Accuracy: 90.32% (calf), 86.67% (cow) |
| (Ding et al., 2022) | Suckling | FD-CNN, YOLOv5 | CSPDarknet53 | FPN | 3 | Precision: 93.6% |
| (Ma et al., 2022) | Motion | CNN (Rexnet 3D) | ResNet101, MobileNetV2, MobileNetV3, ShuffleNetV2, C3D | N/A | 3 | Accuracy: 95% |
| (Gu et al., 2022) | Emergency | Faster R-CNN, YOLOv5, YOLOv4 | CSPDarknet53 | FPN | 3 | Average precision: 95.5% (YOLOv5) |
| (Gan et al., 2022b) | Social | DHRN, STAGCN | Custom-built | N/A | 2 | Precision: 94.21% |
| (Noe et al., 2022) | Lameness | Mask R-CNN | ResNet101 | N/A | 2 | Accuracy: 95.5% |
| (Turner et al., 2022) | Posture, Motion, Feeding | CNN-LSTM, BiLSTM | Custom-built | N/A | 9 | Accuracy: 61.7% using BiLSTM |
| (Shang et al., 2022) | Posture, Motion, Feeding | CNN, YOLOv4 | ShufflenetV2, MobileNetV2, ResNet18, GoogLeNet, MobileNetV3 | N/A | 7, 4 | Accuracy 95.17% using MobileNetv3 |
| (Yu et al., 2022) | Feeding | DRN-YOLO | DRNet | N/A | 3 | mAP: 96.91% |
| (Cheng et al., 2022) | Posture, Drinking, Feeding | YOLOv5 | CSPDarknet53 | FPN | 4 | mAP: 97.4% |
| (Lu et al., 2022) | Posture | YOLOv4 | ESCMobileNetV3 | N/A | 4 | Accuracy: 96.83% |
| (Gan et al., 2022a) | Suckling | CNN | ResNet101 | FPN | 2 | Precision: 95.2% |
| (Uchino and Ohwada, 2021) | Foraging, Motion, Drinking | YOLOv5 | CSPDarknet53 | FPN | 5 | Precision: 95.2% |
| (Jiang et al., 2020) | Posture, Drinking, Feeding, Motion | YOLOv4 | Darknet | N/A | 4 | Accuracy: 97.48% |

Feeding, Foraging, Motion, Mounting, Posture, Social, and Tail-biting. Precision was employed for 11 behaviours, including Drinking, Emergency, Feeding, Foraging, Motion, Mounting, Pecking, Posture, Social, Suckling, and Tail-biting. The Recall was used for ten behaviours, including Drinking, Emergency, Feeding, Foraging, Motion, Mounting, Pecking, Posture, Social, and Suckling. Sensitivity was employed for 07 behaviours, including Drinking, Feeding, Lameness, Motion, Mounting, Nursing, and Tail-biting. Specificity was used for 08 behaviours, including Drinking, Feeding, Lameness, Motion, Mounting, Nursing, Posture, and Tail-biting.

The selection of performance metrics was dependent on the type of DL model. For DL models like CNN and CNN-LSTM, either with pre-defined or custom-built network architecture, metrics such as Accuracy, F1-score, Recall,

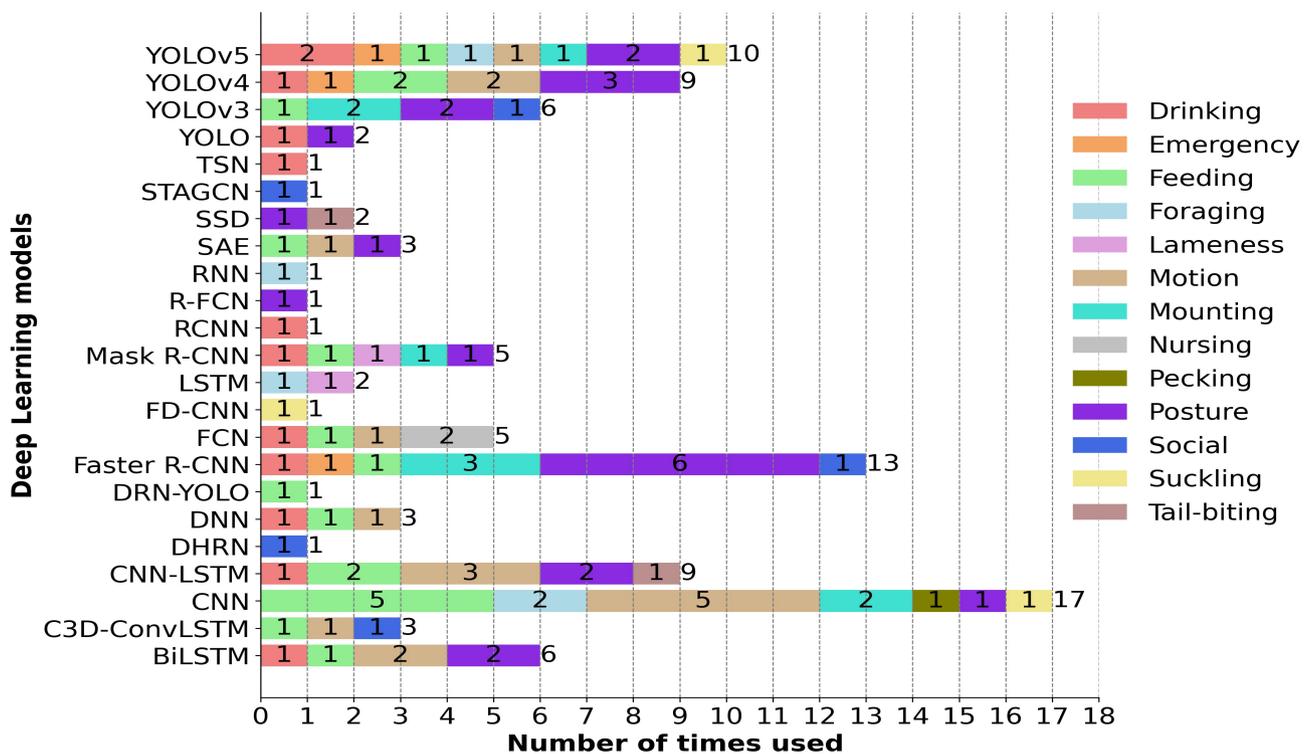

**Figure 5:** DL models reported in the primary studies and their use for each behaviour recognition problem.

Specificity, and Sensitivity were used for evaluation. However, for the DL models such as YOLO, its variants, Faster R-CNN, and Mask R-CNN, metrics such as Precision, mAP, IoU and MCC were used. Notably, the mAP was used to evaluate a DL model in most cases since these models typically use a bounding box strategy for object detection. The mAP computes a score by comparing the ground truth bounding box to the detected box. The higher the score, the more accurate the detection of a DL model. IoU and MCC were the least used performance metrics for group interactive behaviours such as Mounting.

### 3.5. Effective DL models and networks (RQ. 5)

It is difficult to generalise which DL model and network worked best because various classes and performance metrics were used to develop and assess a DL model for a specific behaviour recognition problem in each of the 44 primary studies. For example, for Drinking behaviour recognition, (Wu et al., 2021) used CNN-LSTM, where the model was trained for five classes and used Accuracy as a performance metric. However, (Uchino and Ohwada, 2021) used YOLOv5 with the same number of classes but used Precision as a performance metric. Furthermore, (Alameer et al., 2020a) used YOLO with the same number of classes but used mAP as a performance metric. Therefore, to have a clear overview of the best-performing DL models and networks, this SLR study categorises the results as the most effective DL models and networks based on a particular behaviour recognition problem and overall general performance. Tables 7, 8, and 9 summarise the most effective DL models and networks based on behaviour recognition problems. The results are presented using Accuracy (Table 7), Precision (Table 8), and mAP (Table 9) as performance metrics. While Figure 9 and 10 shows the overall general performance of the top 10 DL models and networks reported in the primary studies regardless of the type of behaviour recognition problems. The highest and lowest reported Accuracy and mAP values are presented for each model and network.

CNN-LSTM, YOLOv5, and YOLO were the most successful DL models for recognising drinking behaviour, whereas the best networks were VGG16, CSPDarknet53, and ResNet-50. CNN and YOLOv5 were the best models for feeding behaviour recognition, whereas GoogLeNet and CSPDarknet53 were the best networks. CNN, YOLOv5, and LSTM were the best models for recognising foraging behaviour, while GoogLeNet and a custom-built network were the best networks. LSTM with a custom-built network provided the best results for lameness recognition. CNN-LSTM, YOLOv5, and CNN were the best models for motion recognition, whereas the best networks were VGG16, CSPDarknet53, and a custom-built network. CNN and YOLOv5 were the best models for recognising mounting behaviour, whereas the best networks were VGG16 and CSPDarknet53. The most successful DL model for nursing behaviour recognition was FCN with the VGG16 network. With a custom-built network, CNN was the best model for pecking recognition. CNN-LSTM and Faster R-CNN were the best models for posture recognition, whereas the best networks were VGG16, ZFNet, and ResNet-50.

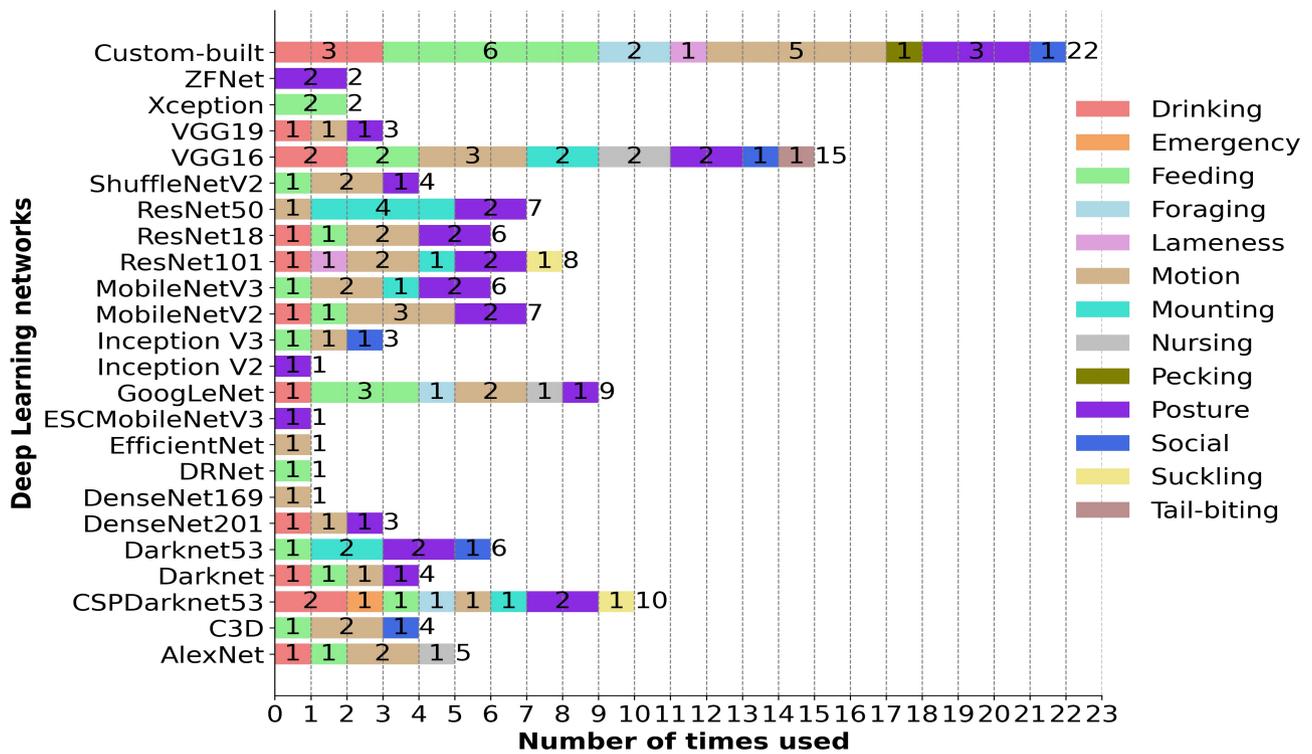

**Figure 6:** DL networks reported in the primary studies and their use for each behaviour recognition problem.

**Table 7**
Most effective DL models and networks based on behaviour recognition problems with Accuracy as a performance metric.

| Behaviour type | Model | Network | Accuracy (%) |
|---|---|---|---|
| Drinking | CNN-LSTM | VGG16 | 98 |
| Feeding | CNN | GoogLeNet | 99.4 |
| Foraging | CNN | GoogLeNet | 99.4 |
| Lameness | LSTM | Custom-built | 96.73 |
| Motion | CNN-LSTM | VGG16 | 98 |
| Mounting | CNN | VGG16 | 98 |
| Nursing | FCN | VGG16 | 96.8 |
| Pecking | CNN | Custom-built | 96.8 |
| Posture | CNN-LSTM | VGG16 | 98 |
| Social | C3D-ConvLSTM | Custom-built | 90.32 |
| Tail-biting | CNN-LSTM | ResNet-50 | 96.35 |

### 3.6. Challenges (RQ. 6)

As DL and AI are still in development, specific challenges are associated with designing and implementing these technologies. Particularly for tasks such as livestock behaviour recognition, there were 13 significant challenges indicated in the primary studies. Figure 11 shows the associated challenges related to the application of DL for livestock behaviour recognition.

**Table 8**
Most effective DL models and networks based on behaviour recognition problems with Precision as a performance metric.

| Behaviour type | Model | Network | Precision (%) |
|---|---|---|---|
| Drinking | YOLOv5 | CSPDarknet53 | 95.2 |
| Emergency | YOLOv5 | CSPDarknet53 | 95.5 |
| Foraging | YOLOv5 | CSPDarknet53 | 95.2 |
| Motion | YOLOv5 | CSPDarknet53 | 95.2 |
| Posture | Faster R-CNN | ZFNet | 95.47 |
| Social | STAGCN | Custom-built | 94.21 |
| Suckling | CNN | ResNet101 | 95.2 |

The most significant of these 13 challenges indicated by the 15.1% of the primary studies was the occlusion and adhesion. Occlusion and adhesion happen when one object in an image obscures the part of another object or when an object comes in contact with another identical object. This often occurs on farms with several animals in a group in an image. Occlusions of several types have been reported; an occlusion can be induced by animals themselves, known as Self-occlusion. Self-occlusion occurs when one part of the body blocks another essential part, inhibiting the features required to recognise patterns in an image for object detection.

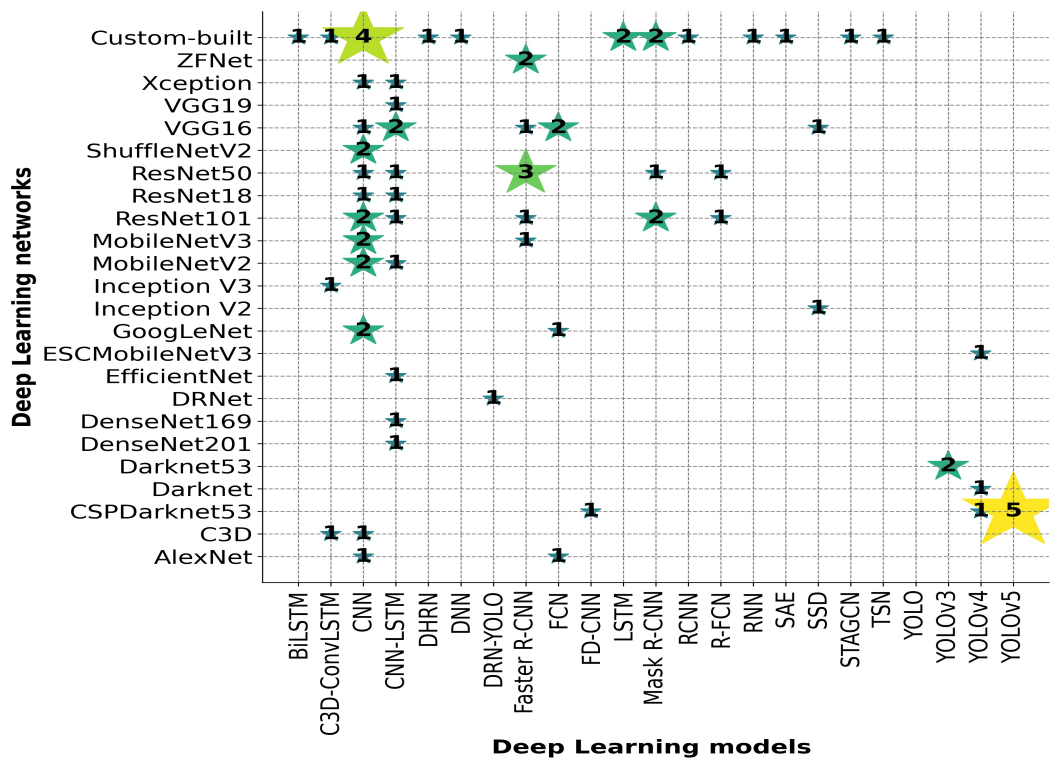

**Figure 7:** The combinations of DL models and networks reported in the primary studies.

**Table 9**
Most effective DL models and networks based on behaviour recognition problems with mAP as a performance metric.

| Behaviour type | Model | Network | mAP (%) |
|---|---|---|---|
| Drinking | YOLO | ResNet-50 | 98.9 |
| Feeding | YOLOv5 | CSPDarknet53 | 97.4 |
| Foraging | LSTM | Custom-built | 86.42 |
| Motion | CNN | Custom-built | 95.4 |
| Mounting | YOLOv5 | CSPDarknet53 | 92.04 |
| Posture | Faster R-CNN | ResNet-50 | 98.9 |
| Social | YOLOv3 | Darknet53 | 85.6 |

Inter-object occlusion can also occur when similar types of objects are in an image. Background occlusion occurs when the object's background under identification mixes with the object.

Data imbalance and complex environments were reported by 13.9% of the primary studies as other prominent challenges. Data imbalance happens when the number of samples for each class is not evenly distributed. Most of the time, there is a large amount of data for one class, referred to as the majority class, and relatively fewer samples for one or two other classes, referred to as the minority classes leading to biased and inaccurate results. On the contrary, the complex environment of livestock farms is another major issue for developing computer vision-based solutions. The environment of these farms often creates complex backgrounds in an image due to the influence of the heat lamps, water and urine stains, manure and complex floor status. The farm environment is variable, and there are partly unpredictable background noises. Interference from postures and surroundings, vast spatial area, and involvement of a substantial number of animals makes the application of DL quite challenging. Another significant challenge reported by 9.3% of the primary studies is related to illumination & weather changes. The different illumination of the scenes, variable light intensity, variation in illumination throughout the day and through different pens or sheds, and data recorded under different light conditions were reported to hinder the development of sustainable DL models.

The lack of availability of adequate data, known as data scarcity, was another challenge reported by 9.3% of the primary studies. Background blending was another challenge reported by 8.14% of the primary studies. Low contrast between animal and background, the animal colour that mixes with the background, makes it hard to differentiate between the background and the object in an image. It was discovered that some animal behaviours were very similar. For instance, there is a resemblance between cow feeding and grass arching behaviour and between piglets' aggressive play and social behaviour. These behavioural similarities were reported by 6.98% of the primary studies. Similarly, 5.81% of the primary studies reported object similarities, the animals with similar colours, shapes, patterns, and sizes in an image as a significant problem. Some other challenges, such as misjudgment among different classes, known as misclassification and missing object detection in videos, were also reported by 4.65% of the primary studies.

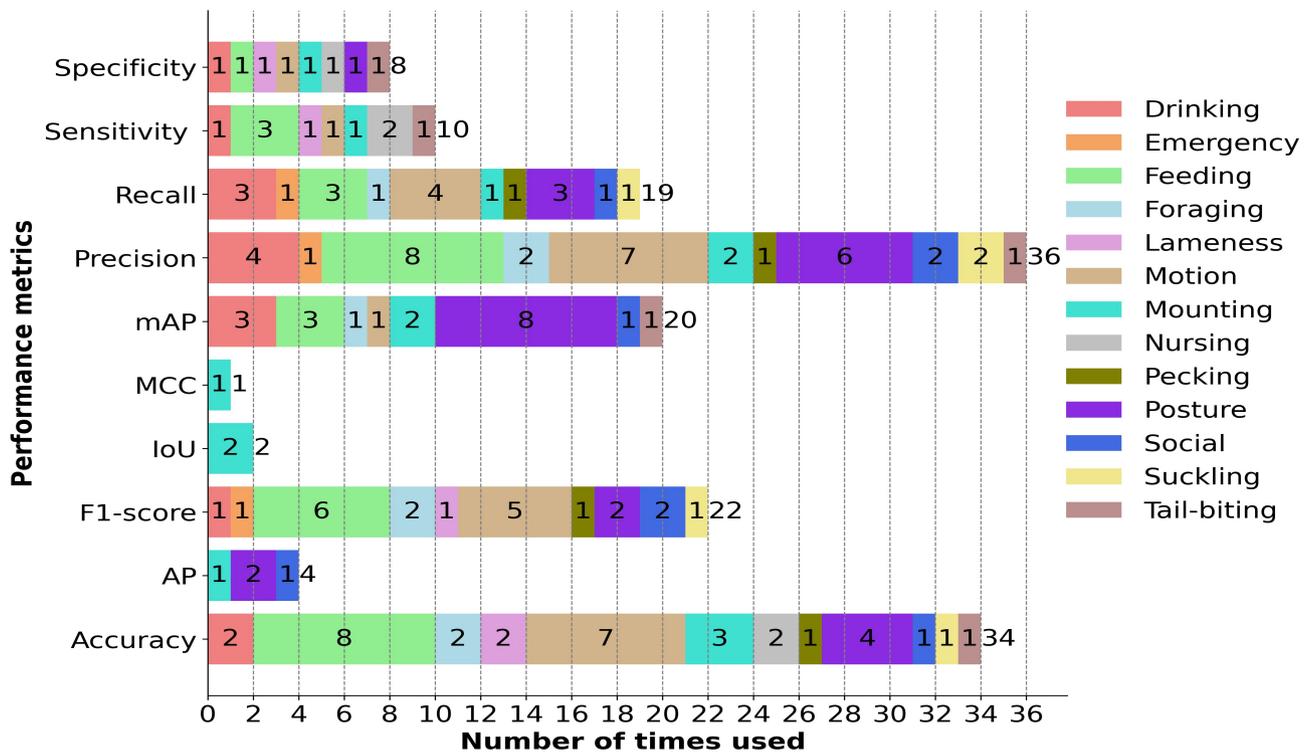

**Figure 8:** The performance metrics used for evaluation in the primary studies.

Although animal behaviour varies widely, there can be significant similarities between various behaviours. Inadequate definitions of behaviours were reported as a barrier in 3.49% of the primary studies to creating more clear-cut and impartial DL models. Another major challenge 2.33% of the studies identified was obtaining robust feature representation in natural scenes, feature extraction, and data collection using high-quality tools.

Occlusion and adhesion, background blending, and challenges related to the complex environment are still open research problems in computer vision. There are several ways to solve these challenges, such as the one using image segmentation presented by (Chilukuri et al., 2022). Most of the challenges reported in the primary studies are directly or indirectly related to the data. The challenge, such as data imbalance, can be solved by recording data with adequate samples for each class. The effect of different illuminations caused by the shadows or weather changes can also be minimised by creating a dataset rich with the samples collected under different lighting and weather conditions. Data scarcity can be reduced by gathering data with good quantity and quality. Challenges related to the object and behaviour similarities and improper behaviour definitions can be solved by defining the behaviours properly so that there is minimum to no confusion amongst different behaviours.

## 4. Discussion

Behaviour recognition is a complex problem that was earlier addressed by installing several physical sensors on the animal's body. These sensors have proven to create distress in animals, negatively impacting their health and well-being. The development of computer vision-based technologies has lately emerged as a more prominent solution than the traditional physical sensor-based approaches.

Although several studies concentrating on the use of vision technologies have yielded some promising results in dealing with the problem of livestock behaviour recognition, there still needs to be an end-to-end approach. Most of the approaches used in the studies are built on frame-by-frame object detection. Object detection models such as YOLO, R-CNNs, and CNNs were used in most primary studies. These models detect objects in a single frame at a time and recognise behaviour by constantly detecting objects in an array of frames (videos).

On the contrary, as the 3D problem, behaviour recognition has been widely discussed in the literature. This is because in detecting a behaviour, there is a need to extract information from a video for a specific interval of time. As previously stated, it can be observed in Table 4 that behaviour is, by definition, time-dependent and consists of a sequence of actions. In the instance of behaviour recognition, time is a critical parameter. In recent years, spatiotemporal analysis has emerged as a much superior approach for action recognition, given that it considers the temporal information in the videos and the objects.

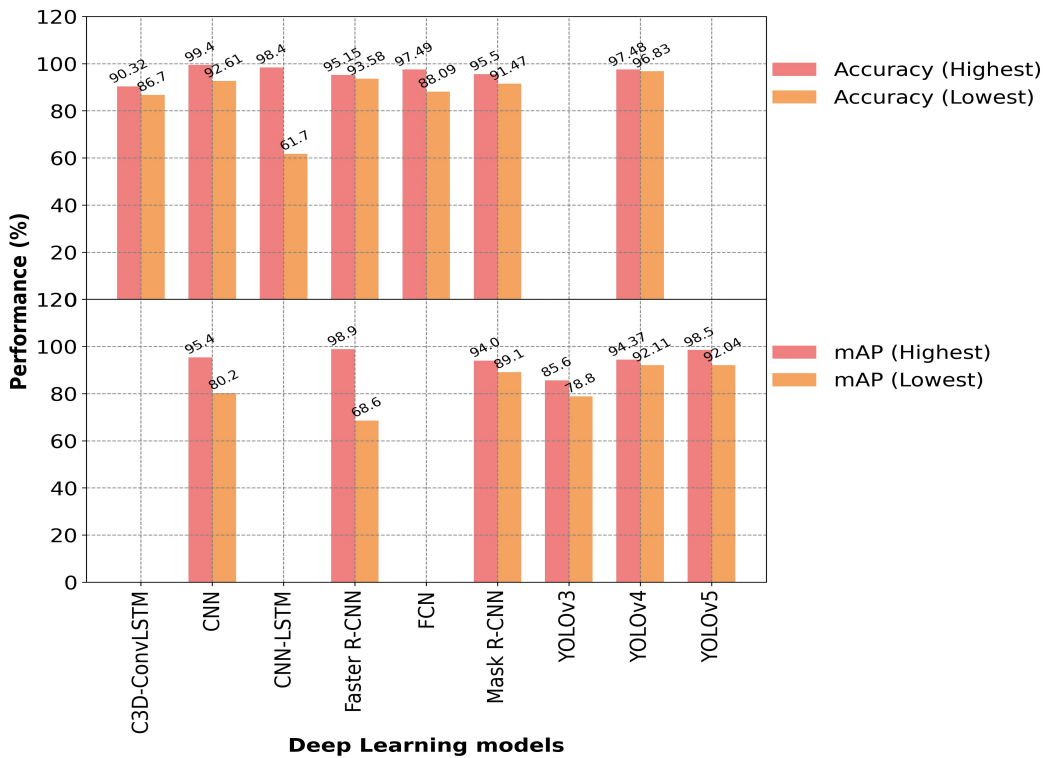

**Figure 9:** The overall general performance of the top 10 DL models.

Although some recent studies have deployed 3D DL models based on spatiotemporal analysis, such as TSN and C3D-ConvLSTM, livestock behaviour recognition is still confined to 2D object detection.

The application of spatiotemporal analysis could enhance the efficiency of DL models. Implementing these strategies requires an adequate amount of data. However, the data is not readily available for livestock behaviour recognition. Only a few publically available datasets can be used to train, test and validate DL models (Ng et al., 2022). Most studies have used custom datasets recorded in several farms under different conditions and are not publicly available. This lack of data availability is a significant bottleneck in developing DL models for livestock behaviour recognition. More public datasets focusing on the behaviours should be generated so that more researchers can contribute to developing these technologies. Furthermore, there are no guidelines for defining behaviours . Multiple definitions were found for a single type of behaviour in different studies. Walking, for example, was described as moving in a standing position, head up and walking, legs moving repeatedly, and cow position changes greatly, and the leg movement for more than 3 secs. Most of these definitions are based on personal observations and do not include expert judgements. A solid uniform set of definitions for each animal behaviour coordinated with veterinary behaviourists can assist in not just collecting tangible data but also in creating DL models more systematically. It will also aid in comparing the effectiveness of various models on the same level. None of the primary studies shared the data used to develop the DL models, It will be also important that wherever possible the data used in the studies is shared with open access to the public to test other possible methods and for the reproducibility of the results.

## 5. Conclusion

This systematic literature review (SLR) investigates the application of deep learning (DL) techniques for livestock behaviour recognition. Accurate recognition of livestock behaviour is crucial for monitoring animal health in controlled agricultural settings. Recent advancements in AI, computer vision, and deep learning have made automated systems that can autonomously identify changes in animal behaviour and underlying health issues possible. This study's findings highlight DL's immense potential in recognising animal behaviours more efficiently than traditional human analysis-based approaches that have been used for decades. However, despite the promising results, DL-based livestock behaviour recognition is still in its early stages of development and faces several challenges. Significant challenges include occlusion and adhesion, complex environments, illumination and weather changes, and background blending, well-known issues in the computer vision community. Research efforts are ongoing to tackle these challenges. Other challenges, such as data imbalance, data scarcity, and data acquisition, can be addressed through tangible strategies for recording high-quality datasets with sufficient information for each behaviour class.

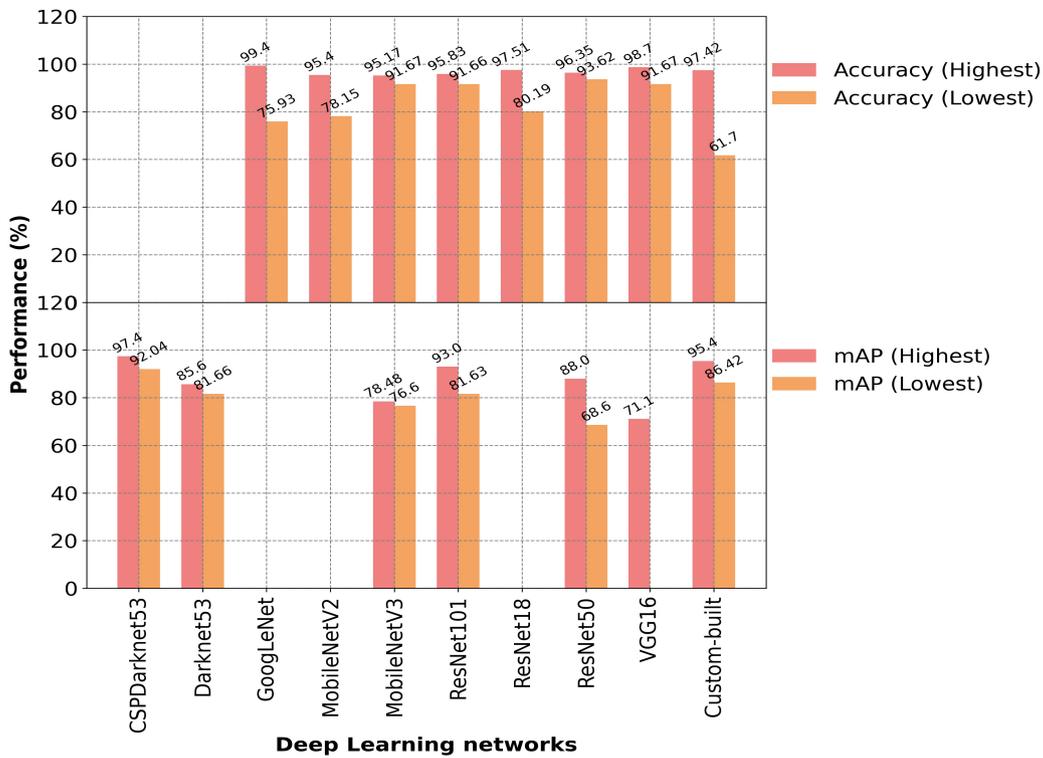

**Figure 10:** The overall general performance of the top 10 DL networks.

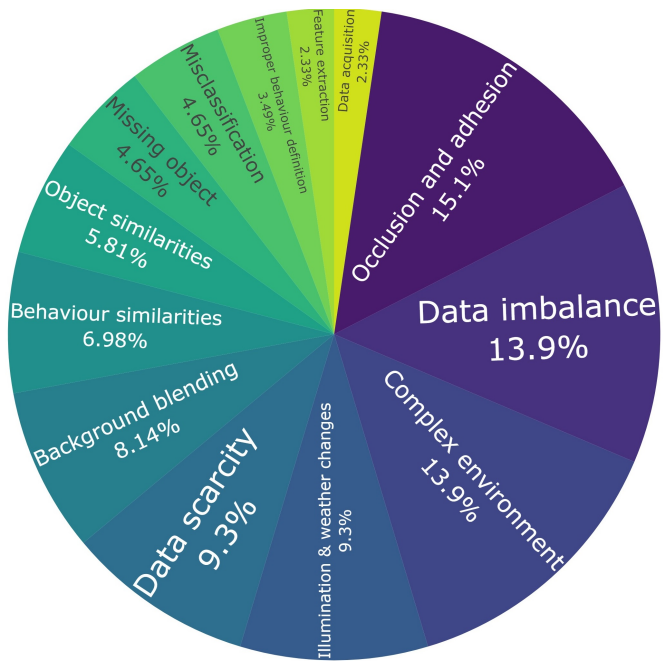

**Figure 11:** Challenges related to the application of DL for livestock behaviour recognition.

Furthermore, no one-size-fits-all DL model can be deemed the best, as researchers have even used various models for similar behaviour recognition problems. However, CNN, CNN-LSTM, Faster R-CNN, and YOLOv5 have shown significant performance compared to other models. Contactless approaches have also been reported to have substantial advantages over contact-based approaches. Additionally, some models, such as C3D Conv-LSTM, have incorporated temporal domain features in addition to spatial domain features. Recent research on human action recognition, which is more advanced than livestock behaviour recognition, has demonstrated that spatiotemporal analysis, considering spatial and temporal features in an image, can yield promising results. This could be an exciting research direction for developing more sustainable DL models for livestock behaviour recognition. Finally, this SLR study is limited to works published between 2012 and October 2022. The articles were selected from the most significant research databases using the criteria outlined in this study; articles from other databases were not included in this review.

## Conflict of Interest

The authors declare that there is no conflict of interest.

## Acknowledgement

This research was financially supported by Innovate UK (Project name: Towards Net Zero Dairy Farming through AI and Machine Vision (DAIRYVISION), Project reference: 107458)

## Declaration of generative AI and AI-assisted technologies in the writing process

During the preparation of this work the author(s) used ChatGPT in order to improve the english language and to correct possible grammatical mistakes. After using this tool/service, the author(s) reviewed and edited the content as needed and take(s) full responsibility for the content of the publication.

## CRediT authorship contribution statement

: Conceptualization: A.R. and T.D.; Data curation: A.R., M.J.H., M.S.R., F.A. and A.K.B.; Formal analysis: A.R., M.J.H., M.S.R., F.A. and A.K.B.; Funding acquisition: T.D.; Investigation: A.R., M.J.H. and F.A.; Methodology: A.R.; Project administration: T.D.; Resources: T.D.; Software: A.R., M.J.H. and F.A.; Supervision: T.D.; Validation: A.R., M.S.R., F.A., A.K.B. and T.D.; Visualization: A.R. and M.S.R.; Writing – original draft: A.R., M.S.R. and A.K.B.; Writing - review editing: A.R., M.S.R., A.K.B. and T.D..